\documentclass[times,twocolumn,final,authoryear]{elsarticle}
\usepackage{jasr}
\usepackage{framed,multirow}

\usepackage{amssymb}
\usepackage{latexsym}
\usepackage{booktabs}
\usepackage{mathtools}

\usepackage{url}
\usepackage{xcolor}
\definecolor{newcolor}{rgb}{.8,.349,.1}

\usepackage[citebordercolor=white]{hyperref}

\journal{Advances in Space Research}

\begin{document}

\verso{Michael Saint-Guillain \textit{et. al.}}

\begin{frontmatter}

\title{Enabling Astronaut Self-Scheduling using a \\Robust Advanced Modelling and Scheduling system: \\an assessment during a Mars analogue mission}

\author[UCL]{Michael Saint-Guillain\corref{cor1}}
\cortext[cor1]{Corresponding author: +32-494-909847; m.stguillain@gmail.com}
% \ead{m.stguillain@gmail.com}
\author[UCL]{Jean Vanderdonckt}
\author[UCL]{Nicolas Burny}

\address[UCL]{Universite catholique de Louvain, Belgium}

\author[ESA]{Vladimir Pletser}
\address[ESA]{European Space Agency (ret.); Blue Abyss, UK}

\author[JPL]{Tiago Vaquero}
\author[JPL]{Steve Chien}
\address[JPL]{Jet Propulsion Laboratory, California Institute of Technology, CA, USA}

\author[SAS]{Alexander Karl}
\address[SAS]{Space Applications Services, Belgium}

\author[AMES]{Jessica Marquez}
\author[AMES]{John Karasinski}
\address[AMES]{NASA Ames Research Center, Moffett Field, CA, USA}

\author[UCL]{Cyril Wain}
\author[UCL]{Audrey Comein}
\author[UCL]{Ignacio S. Casla}
\author[UCL]{Jean Jacobs}
\author[UCL]{Julien Meert}
\author[UCL]{Cheyenne Chamart}
\author[UCL]{Sirga Drouet}
\author[UCL]{Julie Manon}

\received{September 2022}
% \finalform{10 May 2013}
% \accepted{13 May 2013}
% \availableonline{15 May 2013}
% \communicated{S. Sarkar}

\begin{abstract}
%%%
Human long duration exploration missions (LDEMs) raise a number of technological challenges. 
This paper addresses the question of the crew autonomy: as the distances increase, the communication delays and constraints tend to prevent the astronauts from being monitored and supported by a real time ground control. 
Eventually, future planetary missions will necessarily require a form of \textit{astronaut self-scheduling}. 
We study the usage of a computer decision-support tool by a crew of analog astronauts, during a Mars simulation mission conducted at the Mars Desert Research Station (MDRS, Mars Society) in Utah. 
The proposed tool, called \textit{Romie} (\citeauthor{SaintGuillain2022}, \citeyear{SaintGuillain2022}), belongs to the new category of \textit{Robust Advanced Modelling and Scheduling} (RAMS) systems.
It allows the crew members (i) to visually model their scientific objectives and constraints, (ii) to compute near-optimal operational schedules while taking uncertainty into account, (iii) to monitor the execution of past and current activities, and (iv) to modify scientific objectives/constraints w.r.t. unforeseen events and opportunistic science.
In this study, we empirically measure how the astronauts, who are novice planners, perform at using such a tool when self-scheduling under the realistic assumptions of a simulated Martian planetary habitat.
%%%%
\end{abstract}


\end{frontmatter}

%% For linenumbers
%\linenumbers

\section{Introduction}
Past space missions have had very limited experience in human self-scheduling.
In fact, \cite{marquez2019future} states that current human operations, including extravehicular activities (EVAs), are \textit{``carefully choreographed, and rehearsed events, planned to the minute by a large team of EVA engineers, and guided continuously from Earth''} (\citeauthor{bell2012review}, \citeyear{bell2012review}, \citeauthor{miller2015information}, \citeyear{miller2015information}). Activities on the International Space Station (ISS) for example are planned to various detail months and weeks in advance, and transition about two weeks ahead of the planned day into the real-time environment to be reviewed by all teams involved in the activities of that day to allow for further fine tuning in the days before execution. In case of unexpected events requiring an adaptation or re-planning of the day's activities, e.g. equipment failure, or an activity taking considerably longer than anticipated, the decision on how the rest of the day's timeline will be impacted lies with the Flight Director based on inputs by the activity stakeholders and ISS Planners, and then communicated to the crew on board. Activities can move to a different astronaut if there's extra time available, can replace another activity with lower priority, or be moved to another day. The ISS Planners are 24/7 on console and working on the schedules of the coming days and weeks. Today's events impacting tomorrow's timeline will be worked over night while the astronauts are asleep. 
%Current space missions are scheduled in real time. 
%In the ISS, astronauts are in constant communication with teams of experienced planners on Earth, their days being paced to the minute. 
%The future Moon village will most probably be covered by 4G. 
As the distances increase however, the communication delays rapidly become an obstacle to remote real time monitoring and management of operations from Earth.
However, human operations on Mars are expected to be carried out at a faster rate than current rover missions (\citeauthor{mishkin2007human}, \citeyear{mishkin2007human}), which implies new planning strategies and tools that account for latency-impacted interactions (\citeauthor{eppler2013desert}, \citeyear{eppler2013desert}). Current Mars rover missions are commanded by the ground operations team at most once per Martian day, or sol, and operate independently in between such contacts.
In addition, future planetary EVAs are likely to be driven by science (\citeauthor{drake2010human}, \citeyear{drake2010human}, \citeauthor{drake2014human}, \citeyear{drake2014human}), requiring flexible adaptations according to scientific samples. 
In such context, future human space missions will have to enable some degree of crew autonomy and self-scheduling capabilities. 

\begin{figure*}
\centering
\includegraphics[width=1\textwidth]{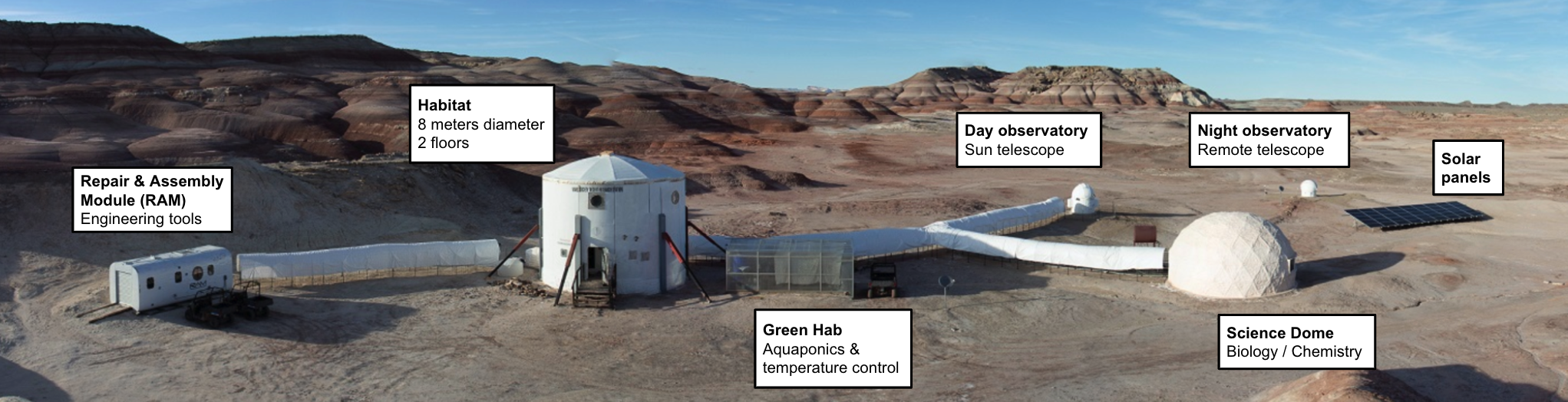}
\caption{\label{fig:mdrs}
The Mars Desert Research Station (MDRS), located in the Utah desert, is a Mars analog planetary habitat (Mars Society).
}
\end{figure*}

The problem of scheduling a set of operations in a constrained context such as the \emph{Mars Desert Research Station} (MDRS, Fig. \ref{fig:mdrs}) is not trivial, even in its classical deterministic version. 
It should be seen as a generalization of the well-known NP-complete  \emph{job-shop scheduling problem} %(\textit{i.e.} conjectured as unsolvable in practice, for real-world sized instances) 
 \cite{lenstra1979computational}, which has the reputation of being one of the most computationally demanding problems \cite{applegate1991computational}. 
\cite{Hall1994} raise on the importance of mission planning, as 25\% of the budget of a space mission may be spent in making these decisions beforehand, citing the Voyager 2 space probe for which the development of the a priori schedule involving around 175 experiments requiring 30 people during six months. 
Nowadays, hardware and techniques have evolved. It is likely that a couple human brains, together with brand new laptops, may suffice in that specific case. Yet, the problems and requirements have evolved too. 
Instead of the single machine Voyager 2, space missions have to deal with teams of astronauts.

\subsection*{Rescheduling on-the-fly: objectives, constraints and opportunistic science.}

Classical space missions are currently scheduled days ahead. Complex decision chains and communication delays prevent schedules from being arbitrarily modified, hence online reoptimization approaches are usually not appropriate.
A human mission on Mars is different. It will necessarily be a long duration mission. The communication delays, in each direction, range from 3 to 22 minutes. Finally, in the current configuration of Mars orbiters, only a few short communication windows with Earth are possible per each Martian day (called a \textit{sol}), with limited data rate (2 Mega bits per second).

In such conditions, any deviation from the original plan must be managed on the fly by the astronauts themselves. 
However, \cite{marquez2021human} demonstrated the fact that astronauts are not good at solving such complex problems by hand. 
This is not surprising. The sheer complexity of space systems means that thousands of constraints must be accounted for in decision making, and balancing of a large number of competing soft objectives must also be considered.  An articulation of the size of this problem space for the Rosetta Orbiter mission science planning is described in \cite{chien_rosetta_aspen_jais_2021} and a future human mission to Mars is likely to be orders of magnitude more complex.   
Furthermore, the astronauts must also be able to adapt their schedules according to new scientific goals and requirements, such as conducting opportunistic science (\textit{e.g.}, recording a dust devil), or even a new scientific project, or unexpected events such as machine breakdowns.
In other words, the human machine team must be able to track evolving scientific objectives and operations constraints to re-optimize activities in an ever changing mission context.  

Whereas our study focused on short-term scheduling, it is worth noting that there are several perspectives, ranging from tactical planning (short term) to strategic planning (long-term). In other words, various levels of granularity: mission objective vs instrument-specific procedures. 
Some may easily be transferred into a crew autonomy (like scheduling a repair), others will have to reside on the ground segment, as they need for instance a scientific debate on reshaping scientific objectives along the mission.

\subsection*{The impact of uncertainty}

At the MDRS, computing an optimal schedule becomes significantly less attractive as problem data, such as the manipulation time of experiments, are different from their predicted values. In a constrained environment with shared resources and devices, such deviations can propagate to the remaining operations, eventually leading to global infeasibility.
The purpose of this paper is to investigate, based on the real case study of a Mars analog mission, the impact of stochastic robust modeling against a classical deterministic approach on the reliability of a priori mission planning. 
% \begin{figure}
% \centering
% \includegraphics[width=1\columnwidth]{figures/MDRS_EVA.png}
% \caption{\label{fig:mdrs_eva}
% Left: the Mars Desert Research Station (MDRS) in the Utah desert, U.S, is a Mars analog planetary habitat. Right: extra-vehicular activity on field operations (EVA).
% }
% \end{figure}

% \subsubsection*{Illustrative example}

\begin{figure}
\centering
\includegraphics[width=.9\columnwidth]{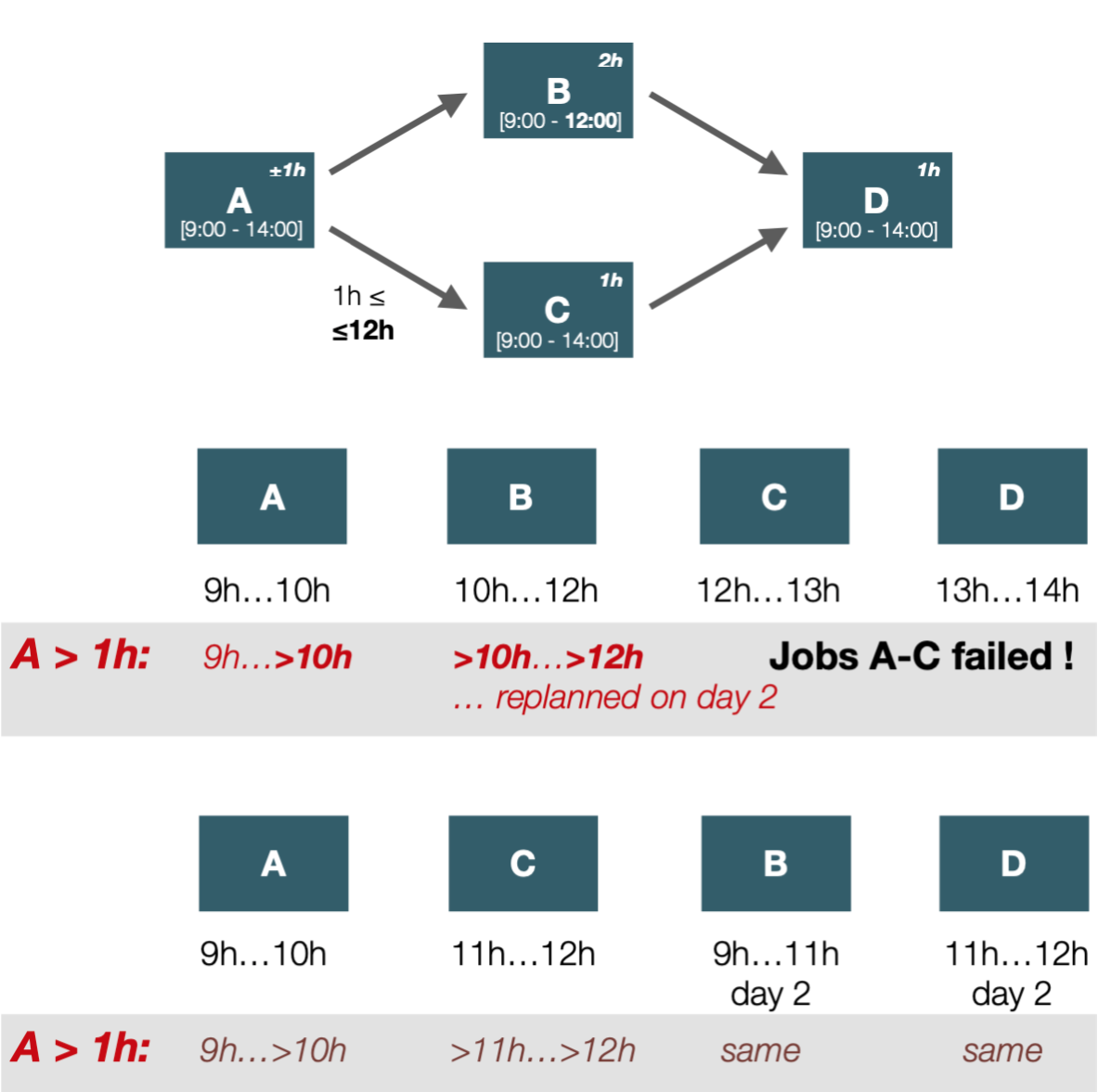}
\caption{Illustrative example with four tasks: \textit{A}, \textit{B}, \textit{C}, \textit{D}, to be scheduled on the same line. Each task has a processing time of 1 or 2 hours, and a time window spanning either the entire work day (9 am to 2 pm) or part of it (9 am to 12 pm or 12 noon). 
Tasks \textit{B} and \textit{C} require \textit{A} to be completed, and \textit{D} requires both \textit{B} and \textit{C} to be completed, before being started.
Task \textit{C} must wait at least one hour after completion of \textit{A} to start, and must be completed during the same day. }
\label{fig:example}
\end{figure}

Consider the simple project depicted in Figure \ref{fig:example} (top).
Suppose all tasks have to be scheduled on the resource, then one must necessarily begin with $A$ and end with activity $D$. 
There are only two valid schedules, shown in Figure \ref{fig:example} (bottom).
In fact, schedule $(A,B,C,D)$ looks much more efficient, as all tasks are completed on the first day. On the contrary, schedule $(A,C,B,D)$ requires an additional day.
However, this is \textit{only true on the paper}, when everything is predictable. 
If you account for (temporal) uncertainty, then the story is different. 
If operation $A$ lasts for more than 1 hour, schedule $(A,B,C,D)$ is \textit{not valid anymore}: $B$ will have to be resumed or rescheduled on day 2 (an additional day, that was not expected!). When starting $C$, we realize the worst: it actually requires to be processed the same day as $A$. Mission failed. 
Remark that if the true average processing time of $A$ is 1 hour, then this scenario happens with at least 50\% probability.
On the contrary, (almost) whatever happens to $A$, under schedule $(A,C,B,D)$ everything goes fine. 
This schedule is said to be \textit{robust}. Its success probability is simply 1 minus the probability that $A$ exceeds four hours (which we assume to be fairly unlikely).

\subsection*{Application and contributions.}
Contrary to current space missions, in which astronaut operations are paced to the minute and supervised in real time by experienced planners on Earth, future long duration exploration missions (LDEMs) will necessarily involve a certain level of autonomy, or \textit{self-scheduling}.

In this paper, we study the ability of astronauts, being novice planners, to organize themselves their operations. 
The problem at stake being intrinsically complex, a decision-support system is provided. 
We hence measure, report and analyse how efficiently the astronauts are exploiting such technology to pursue their scientific objectives, \textit{autonomously}, and during a real Mars analog mission carried out at the MDRS.
We demonstrate the usability and usefulness of a decision support system such as \textit{Romie}, and compare its user experience with an existing one (Minerva, Nasa), in terms of human-machine interactions.

\section{Operations management software systems \label{sec:literature}}

% There are many scheduling tools on the market. In a recent study, \cite{abramov2016analysis} describe some of the most common ones (MS Project, Primavera, Artemis, \textit{etc}).  
% However, one should pay attention to a fundamental difference between software systems under the very large denomination of "\textit{scheduling (or planning) tools}", and the so-called "\textit{advanced planning and scheduling}" (APS) tools (\citeauthor{stadtler2015supply}, \citeyear{stadtler2015supply}). 
% When called APS, in addition to the ability of visualizing and manipulating schedules, the tool should come with an automated scheduling and optimization engine, able to generate solutions (\textit{i.e.} schedules) based on a set of pre-specified operations and constraints (usually hidden to the user!). 
% On the contrary, the vast majority of non-APS tools, such as MS Project, are mostly complex Gantt chart manipulation software systems. In other words, the user is still left with the initial problem of providing a scheduling solution, that is, the ordered sequences of tasks, each being assigned resources and scheduled times. 
% In what follows, we focus on existing systems that provide both the modeling and the optimization capabilities.

Existing systems usually fall into \textit{a)} being specifically designed for a particular application/mission or operational context, or \textit{b)} not having a generic, integrated  optimization system to generate robust schedules (from a probabilistic point of view).
Instead, the \textit{Romie} RAMS system % As a follow up to my earlier comment, a simple description of the \textit{Romie} RAMS system should be given here or at least a reference should be given  where this information can be found (VP)
is used in this study. 
Compared to classical frameworks, a robust advanced modelling and scheduling (RAMS) system such as \textit{Romie} provides the following \textit{technological innovations}:
\begin{enumerate}
    \item Graphical problem modelling. The user is able to graphically draw and manipulate the structure and constraints of its scheduling problem, including stochastic models for task durations. 
    \item Optimization under uncertainty. An optimization engine allows the user to generate, or adapt existing schedules, in a way that produces schedules robust w.r.t. uncertainty. 
\end{enumerate}

 \subsection{Planning and Scheduling in Space \label{sec:aspen}}

The first planning and scheduling tools for space missions were dedicated software systems, specialized to specific application domains. 
% For instance, \cite{muscettola1993hsts} developed the HSTS framework for short term scheduling 
\cite{johnston1994intelligent} described the SPIKE system, a general framework for scheduling, developed by the Space Telescope Science Institute for NASA's Hubble Space Telescope. 
Other examples of aerospace scheduling tools and applications are: \cite{chien1999automating}, developed for scheduling the operations of a particular shuttle science payload (DATA-CHASER) with primary focus on solar observation; \cite{jonsson2000planning} for the Deep Space One mission; \cite{ai2004mapgen} for the Mars Exploration Rover mission; \cite{chien2005using} for NASA's Earth Observing One Spacecraft; and \cite{cesta2007mexar2} for the Mars-Express mission.
\cite{chien2012generalized} provides a detailed survey on (semi-)automated planning \& scheduling systems developed for space applications. 

As the need for more generic approaches to support multiple mission and multiple domains increased, a planning/scheduling C++ library has been proposed: ASPEN (\citeauthor{fukunaga1997towards}, \citeyear{fukunaga1997towards}, \citeauthor{rabideau1999iterative}, \citeyear{rabideau1999iterative}, \citeauthor{chien2000aspen}, \citeyear{chien2000aspen}). At that time, ASPEN provided the elements that were commonly found in existing complex planning and scheduling systems, for example for generating operation schedules for the Rosetta orbiter \cite{chien_rosetta_aspen_jais_2021}. 
In 2009, ESA's Advanced Planning and Scheduling Initiative (APSI) aimed at developing a general software framework for supporting development of AI planning and scheduling prototypes, for various types of space missions. The APSI is described in \cite{steel2009advanced}.

Presented in \cite{yelamanchili2020ground}, the \textit{Copilot} system for Mars 2020 Rover mission does have a modelling system called COCPIT, and a planner, but it is specifically designed for that mission.  This ground automated planning system is intended for use with an onboard planner in preparation for deployment \cite{rabideau-benowitz-IWPSS-2017, agrawal_flexible_robots_2021, agrawal_switch_groups_jais_2021}.  Of particular relevance to this work is the Copilot ground scheduler with explanation capability \cite{agrawal_m2020_xaip} and Monte Carlo variation of execution to set parameters for onboard rescheduling \cite{chi_icaps2019_optimizing}.  Again, these systems are fairly tailored to the specifics of the Perseverance rover mission.

A key differentiation in space missions is human surface missions versus automated orbital, flyby and other space mission modalities.  Flyby and orbital missions can be well predicted, enabling pre-planning of observation campaigns (often days or weeks in advance) and executed (excepting fault protection) primarily open loop. Some exceptions to this generalization over non surface missions exist. For example VML was used onboard Spitzer to enable it to handle variable execution time or failure to acquire guide stars for observations. JWST has a similar capability.  Some examples of such predictable missions that have used automated scheduling include (non exhaustive list)  MAMM, Orbital Express, Hubble, Spitzer, Earth Observing One, and Rosetta Orbiter to name a few (more are described in \cite{chien2012generalized}).  In contrast, surface missions, especially those involving astronauts (such as human exploration of Mars), involves more intimate interaction with the environment and are therefore harder to predict.  Previous lander and rover missions have encountered challenges in variability of action duration (e.g. driving), challenges in physical manipulation (e.g. placing measurements, drilling and coring, ...) which might mean activity failure.  Such challenges in unpredictability of execution are strong motivation for the capability of any human-machine joint system to be able to continuously replan in light of such occurrences (see \cite{gaines-doran-justice-et-al-2016} for an excellent study of such challenges for the Mars Science Laboratory Mars Rover Mission and \cite{gaines2020jfr} for work at increasing the ability of future Mars Rovers to autonomously redirect their activities in such situations.).  Note that this autonomous handling of uncertainty is at a premium for future missions to explore unknown environments such as the Europa Lander Mission Concept \cite{europa-lander-uncertainty-icaps2022}.

\subsection{Human Self-Scheduling in Space \label{sec:self-scheduling-space}}

In \cite{deans2017minerva}, a suite of software tools called Minerva is proposed in order to support operations planning and execution. Minerva and its components (xGPS, Playbook, SEXTANT) have been tested during several planetary and space simulation missions, including the BASALT research program (described in \citeauthor{brady2019strategic}, \citeyear{brady2019strategic}) and four analog missions at NEEMO (\citeauthor{chappell2017integration}, \citeyear{chappell2017integration}, \citeauthor{marquez2017increasing},  \citeyear{marquez2017increasing}). 
Compared to Minerva, the key differences of our proposed tool \textit{Romie}, in terms of functionalities, rely on the modelling interface and the scheduling optimization engine, which enable strategical a priori planning. In addition, the optimization is conducted while taking uncertainty into account. The Minerva suite is rather focused on tactical planning, including geospatial planning, which allows crew path planning and coordination using satellite maps. The strategical planning is assumed to be performed before the start of the mission, and is therefore not covered by the Minerva suite. 
However, even when a predefined schedule is provided prior to the start of the operations, it is very likely that the schedule will require online modifications as the operations go. \cite{marquez2021human} showed the limits of human self-scheduling when operators must solve and adapt the planning manually while taking hard constraints into account (not even thinking about uncertainty).
By providing both a way to adapt the model and solve it using an embedded optimization engine, \textit{Romie} is complementary to Minerva.

% (\citeauthor{miller2017decision}, \citeyear{miller2017decision})

\begin{figure}
\centering
%    \vspace{-8pt}
\includegraphics[width=.9\columnwidth]{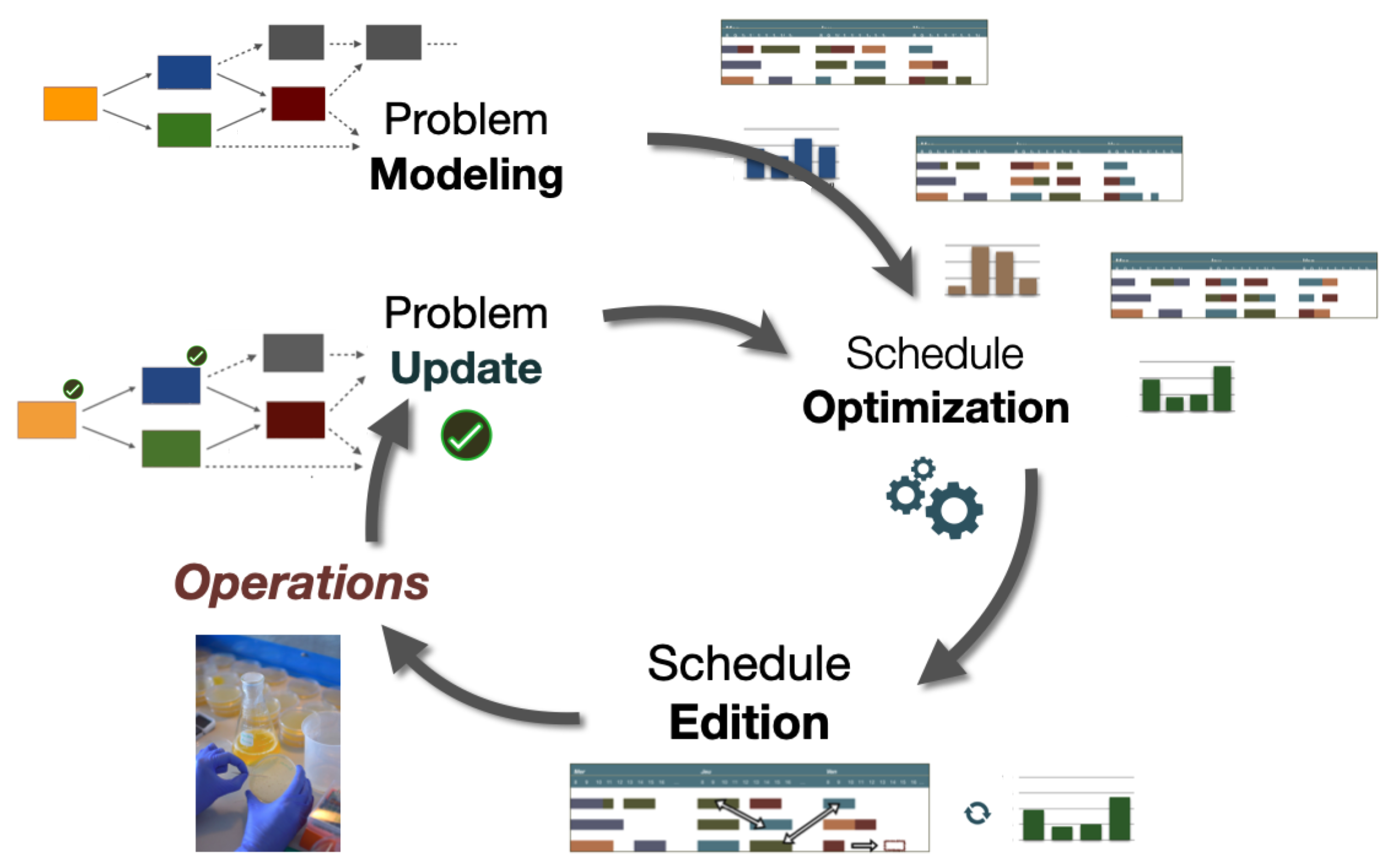}
% \vspace{-16pt}
\caption{Overview of the \textit{Romie} modeling and scheduling system.}
\label{fig:lifecycle}
\end{figure}

\subsection{Romie}
Recall the two technological innovations of Romie, presented in the beginning of this section: \textbf{a)} domain-independent graphical modelling (and scheduling) interface and \textbf{b)} optimization under uncertainty. 
Unlike all existing tools, both modelling and modifying the problem is now made accessible to the end-user, which is critical for a reliable self-scheduling.
Up to our knowledge, the MapGen tool presented in \cite{ai2004mapgen} was one of the very first tools to propose a visual constraints editor. However, the latter was not generic, but specific to its application case, the NASA's MER mission.

\textit{Romie} is the first scheduling tool to propose an integrated robust (\textit{i.e.} under uncertainty) optimization engine. Having more robust (\textit{i.e.} reliable) schedules, the end users are more likely to avoid last minute rescheduling. \textit{Eventually, what-if analysis, as well as sensitivity analysis, become less relevant}: by considering the uncertainties related to task execution, the solutions are optimized following directly the \textit{expected values} of the chosen key performance indicators (KPIs).

We believe that both \textbf{a)} and \textbf{b)} provide significantly more autonomy to the end users, whom remain otherwise highly dependent of planning and scheduling experts. Based on the theoretical foundations defined in \cite{SaintGuillain2021}, the empirical contribution of point \textbf{b} has been extensively validated by \cite{SaintGuillain2019}, \cite{SaintGuillain2022biotech} and \cite{SaintGuillain2022}.
Testing the ability of the non-experts end-users to actually "self-schedule" using \textbf{a)} is the main goal of this study. 

Figure \ref{fig:lifecycle} depicts the key functionalities of our system:
\begin{itemize}
    \item \textit{Graphical modeling} of the problem at stake, in its own operational context: human and physical resources, operational constraints, key performance indicators (KPIs), execution uncertainties.
    \item \textit{Robust scheduling}: the optimization engine takes the time uncertainty on each task's duration into consideration, using modified-PERT distributions, yielding schedules with high probability of success. 
    \item \textit{KPI-guided optimization}: The schedules are optimized while pursuing (a combination of) \textit{various KPIs}, including success probability, expected cost, expected quality, and even operator wellness.
    \item \textit{Operations update} and \textit{online reoptimization}. The system knows the difference between past and future operations. As the schedules can be extensively modified by hand, in particular in the past (but also in the future), reoptimization on future decisions can be performed based on what actually happened in the past.
\end{itemize}

%%============
\section{The M.A.R.S. UCLouvain 2022 mission \label{sec:mission}}

% \subsection{Before the simulation}
Our study on astronaut self-scheduling is driven by the scientific research projects to be carried out by the crew members in the context of the simulation. Before the actual beginning of the mission, the selected projects have been modelled in the \textit{Romie} system, and provisional schedules have been designed. 
In what follows, the different projects are described. Thereafter, their modelling and a priori scheduling is analyzed, from the user's point of view.

\subsection{Experimental plan}

Our study aimed at answering the following questions: 
how long does it take for a novice user before setting up correct schedules (on-boarding time), and 
are our astronauts all able to adapt their scientific objectives as the operations evolve?
We tackled these questions by focusing on the temporal evolution of these following two complementary KPIs: system usability and user experience. 
The ISO-9241-210 standard \cite{ISO9241-210:2019} defines the usability as \textit{the extent to which a system, product or service can be used by specified users to achieve specified goals with effectiveness, efficiency and satisfaction in a specified context of use}. 
User Experience is defined by the same standard as \textit{the user’s perceptions and responses that result from the use and/or anticipated use of a system, product or service} and is generally understood as inherently dynamic, given the ever-changing internal and emotional state of a person and differences in the circumstances during and after an interaction with a product \cite{Vermeeren:2010}.

Several scientific research projects were conducted at the MDRS. Each project was carried on in place, by either one or two astronauts. Some projects (such as health projects) involve the participation of all the crew members.
Yet, these projects were designed and prepared months ahead. During that period, preliminary experiments were conducted on \textit{Romie}, providing first results on the system's usage by the astronauts, in offline (supervised) conditions. 
The actual \textit{M.A.R.S. UCLouvain 2022} mission period, which lasted 12 days on field at the MDRS (see Figure \ref{fig:mdrs}), constituted the main material of this study. 
Day after day, each crew member used the \textit{Romie} system to monitor and update their operations. 
%Measures using different techniques will be made each time the astronauts use the system, in order to record and further analyze how well they succeed at self-scheduling.

\subsection{The Mars Desert Research Station}
\begin{figure}
\centering
\includegraphics[width=\columnwidth]{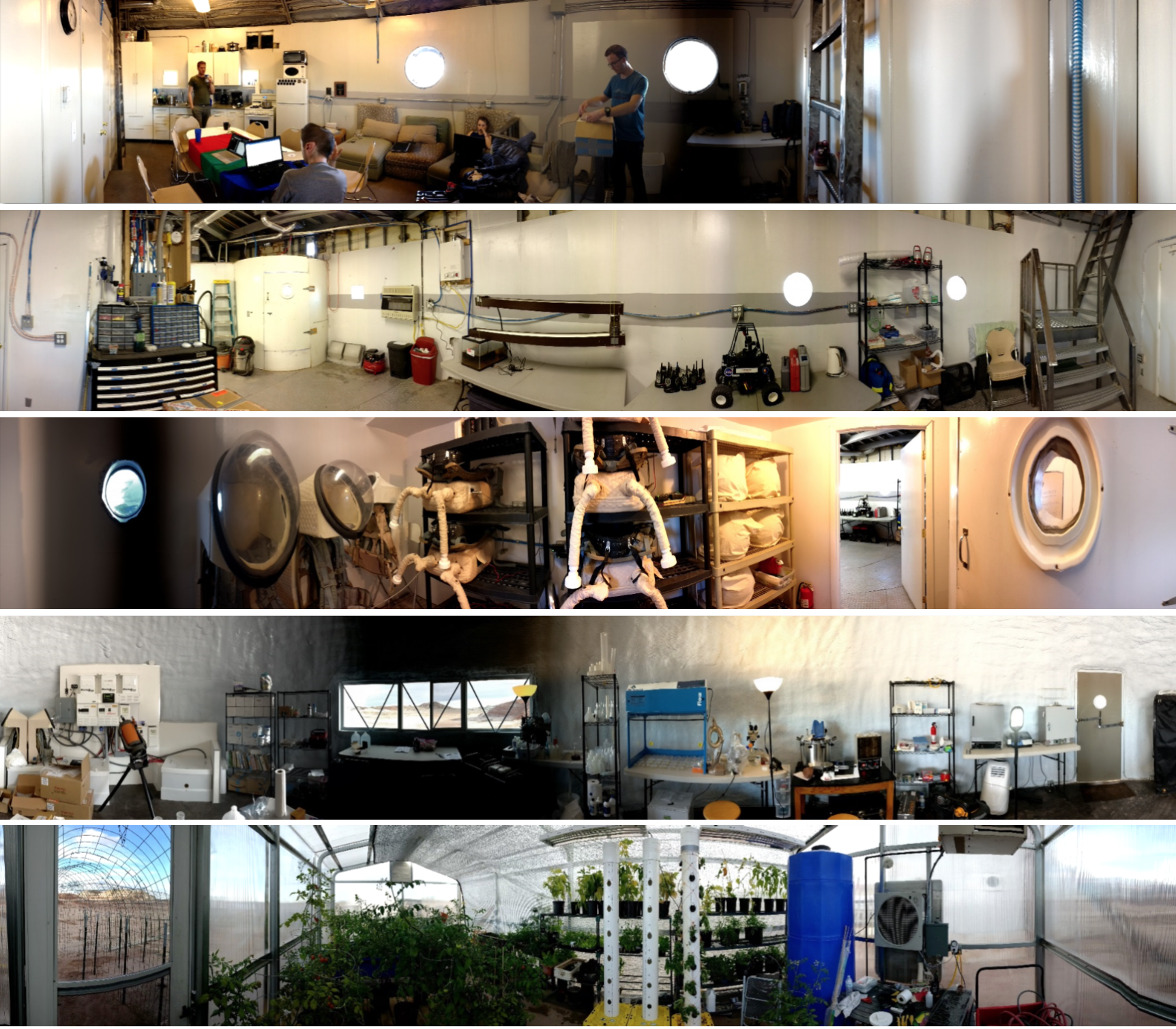}
\caption{Panoramic pictures of some of the Mars Desert Research Station (MDRS) elements, from inside. From top to bottom: upper deck, lower deck, EVA preparation room, science dome, green hab.}
\label{fig:MDRS_pano}
\end{figure}

The MDRS in the desert of Utah has been in operation since 2002 from November through April every year. The geologic features of the surrounding Jurassic–Cretaceous terrain also make the desert environment seem Mars-like to crew members. The MDRS habitat itself is a vertical cylindrical structure of approximately 8 m diameter and 6 m high, composed of two floors. The ground floor (lower deck) includes a front door airlock used for simulated EVA, an EVA preparation room, a large room used as a laboratory for geology and biology activities, a small engineering workshop area, a second back door airlock for engineering activities, a small bathroom and a toilet, three small windows, and a stair leading to the first floor. The first floor (upper deck) includes a common area or living room with a central table, a wall-attached circular computer/electronic table, a kitchen corner, six small bedrooms, and a loft on top of the small bedrooms. 
Some panoramic pictures from the inside are provided in Figure \ref{fig:MDRS_pano}.

\subsubsection{A typical day on Mars}
The day-to-day operations at the MDRS is as follows. 
The crew wakes up at 7:30. Then directly follows a twenty minute morning sport session, before having breakfast, which is typically the right moment for daily medical examinations. 

Extra-vehicular activities (EVAs) take place during the morning. 
Between three and five crew members get prepared for the daily EVA. That takes roughly one hour, during which the crew members that are not participating in the EVA help the others to don their spacesuits, and parameterize the communication devices. 
The EVA should start no later than 9:30, as it must necessarily be ended before 12, which provides roughly two hours to reach all the EVA objectives. The crew members that remain inside MDRS stay in permanent contact with the EVA party, while performing the daily chores. 

Scientific activities then take place every day from 1:30pm to 6pm. The crew members work on separate places, depending on their research field: the crew botanist stays in the green hab, biologists and chemists in the science dome, the astronomer takes pictures of the sun in the day observatory, engineers work in the repair \& assembly module (RAM), \ldots
During each afternoon, the crew members, one by one, use the RAMS system to monitor and schedule their operations. 
Therefore, each crew member uses the \textit{Romie} system \textit{–and answer the questions and exercises defined in the scope of our study–} once a day, for approximately 30 to 60 minutes.

From 6pm, all the crew members would generally interrupt their activities, in order to prepare for the daily communication window, from 7 to 8pm. 
At exactly 7pm all the specific reports are sent to ground control: engineering, medical, green hab, EVA and EVA request, journalist, and commander report. While sharing the diner, the entire crew remains available to answer questions on these reports. The remaining of the evening constitutes a privileged, necessary moment for socialising.

\begin{figure*}
\centering
\includegraphics[width=\textwidth]{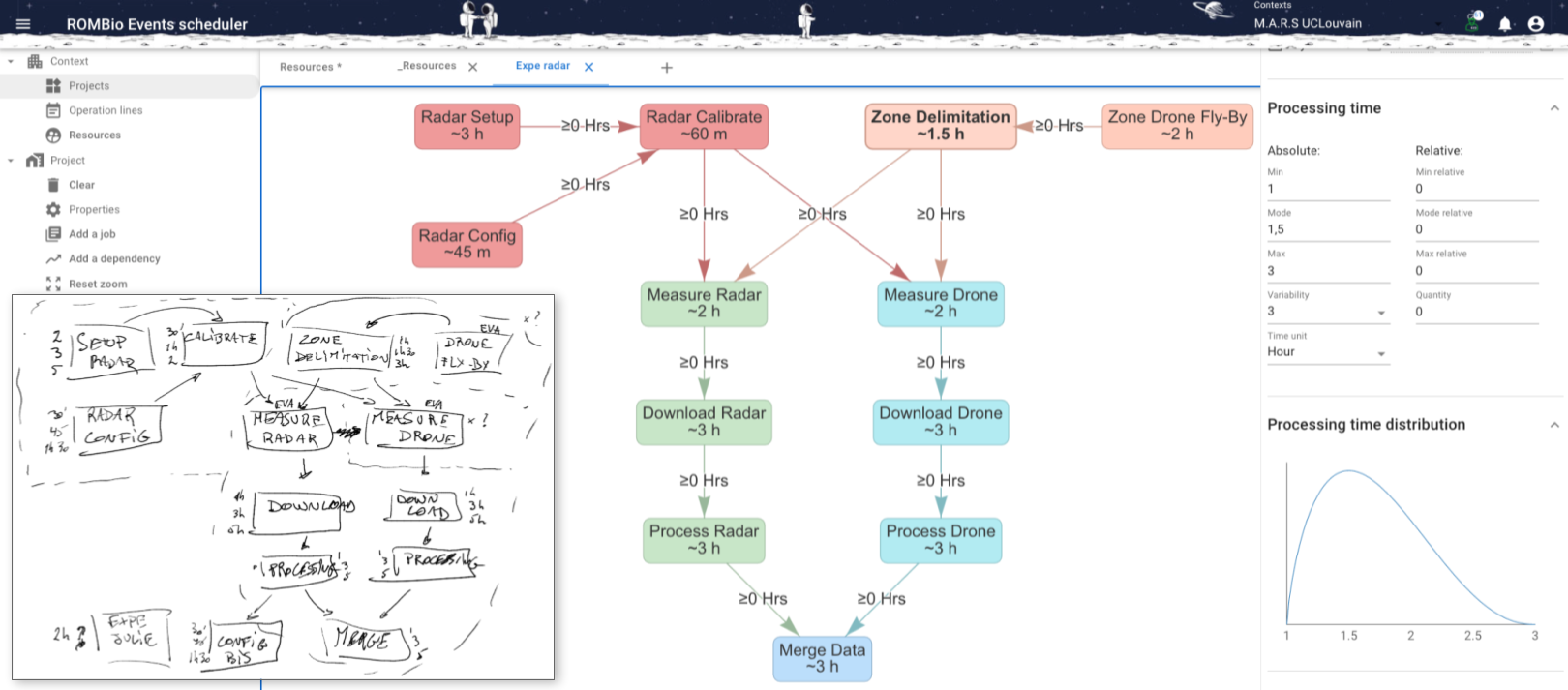}
\caption{Modelling research project \textit{``Soil dielectric 3D map''} within \textit{Romie}'s graphical interface. Boxes represent activities. Arrows represent (temporal) precedence constraints. Here the \textit{Zone Delimitation} activity is selected, showing the parameters (right panel) that define its temporal uncertainty: min, mode, max. This activity requires \textit{Zone Drone Flyby} to be completed before, and is a prerequisite to both activities \textit{Measure Radar} and \textit{Measure Drone}. Bottom left: original hand-drawn sketch.}
\label{fig:model_cyril}
\end{figure*}

\subsubsection{Time-eaters at MDRS}
Previous studies (\citeauthor{pletser2009european}, \citeyear{pletser2009european}; \citeauthor{boche2009mdrs}, \citeyear{boche2009mdrs}; \citeauthor{boche2009human}, \citeyear{boche2009human}; \citeauthor{pletser2010crew}, \citeyear{pletser2010crew}; \citeauthor{pletser2010mars}, \citeyear{pletser2010mars}; \citeauthor{thiel2011human}, \citeyear{thiel2011human}; \citeauthor{pletser2011european}, \citeyear{pletser2011european}) have shown that there are many 'time-eaters' in a day at the MDRS during a simulated Mars stay mission. 
Table \ref{tab:time-eaters} reports the measured average unproductive time of the 76th rotation in 2009.
A hypothetical average crew member would have only approximately 7 hours left for scientific work. Even if this average estimation is very crude, it shows that the remaining average time for scientific work is significantly low and that a lot of time is spent on unproductive tasks, chores and maintenance. 
These data are consistent with previous findings for crew 5 in 2002 (\citeauthor{clancey2006participant}, \citeyear{clancey2006participant}) in terms of activity duration. 
Furthermore, the time-sharing of occupation of the ground floor laboratories between geologists and biologists was always difficult to establish and necessitated a lot of good-will of all parties. Both scientists' teams had different needs and expectations, e.g., biologists need a clean, pristine and well-lit environment to analyse and process soil samples, while geologists need a darker environment to handle, manipulate, crush and process samples with instruments, often generating dust and noise. 
In order to share the use of the single room laboratory, both groups of scientists had to work during the night alternately. Traffic of crew members through common and scientific areas was another point of study as it created also interruption of science work by engineers to assess the Hab systems and by other crew members to access stowage areas, etc. 
It highlights the importance of having a proper and performing dynamic planning tool that can be used on the spot and on the run, fine tuning and adapting an already agreed day planning, for example, either in the evening at dinner or in the morning during breakfast.

\begin{table}[]
\centering \footnotesize
\begin{tabular}{ccccccc}
Breakfast & Lunch & Dinner & Chores & Maint. & Evening  & Sleep  \\ \hline
44m       & 48m   & 57m    & 3h08m  & 1h23m  & 1h35m & 8h26m \\
±02m      & ±02m  & ±01m   & ±18m   & ±10m   & ±13m  & ±07m  
\end{tabular}
\caption{Average durations with standard errors for an average crew member. Note: Based on measured activity durations of Crew 76 (see \citeauthor{pletser2011european}, \citeyear{pletser2011european}); h= hours; m: minutes. Maint.: maintenance. Evening: evening common activities. It sums up to 17 hours 01 minute, ±53 minutes.
\label{tab:time-eaters}
}
\end{table}

Several recommendations were made to improve the design in order to optimize the traffic and to decrease the time spent unproductively from a scientific point of view.
Yet, the ``time-eaters'' cannot be completely avoided. The system proposed in the current study comes in addition to these recommendations, as we investigate an AI based decision system to optimise productivity while leveraging unpredictable time deviations.

Naturally, being a non-scientific work, using the \textit{Romie} system should also be considered as a time eater. 
Its per-person usage duration (half an hour to one hour per day) should probably be reduced to be usable in an actual mission context. Another way to save time would be to assign the responsibility to only one astronaut, thus being the crew planner, to manage the global mission schedule by using the \textit{Romie} system. However, our study required several end users.

\subsection{Research projects}

Each analogue astronaut has her/his own research objectives for the mission. In fact, each astronaut (experimenters) prepared one different research project to be carried out at MDRS. There are thus eight research projects, from eight different fields such as biology, botanic, engineering, astronomy or medicine:

\textbf{Soil dielectric 3D mapping:} Using a ground penetrating radar, installed on a vehicle, the dielectric properties of the soil surrounding the station are measured and projected on a 3D map, constructed by photogrammetry using a drone. Such a map could be exploited to optimize future irrigation systems. 
% This project is lead by \textit{Cyril Wain} (crew commander).
\textbf{3D printing:} This experiment exploits 3D printing scaffolds in bio-ink to seed stem cells, and performs mechanical stress-strain tests on the resulting micro-architecture.
% This project is lead by \textit{Ignacio Sanchez Casla} (crew astronomer).
\textbf{Sleeping hypnosis:} This project tests an hypnosis technique, used in medicine before falling asleep, to help the astronauts having better, deeper sleeps. 
% This project is lead by \textit{Julien Meert} (crew engineer).
\textbf{ExFix:} 
Accidents and injuries on Mars are dangerous. \cite{Manon:2023} study a low-cost external fixator to stabilize broken bones, which remains accessible, fast and easily achievable by any astronaut without surgical training.
% This project is lead by \textit{Julie Manon} (health \& safety officer).
\textbf{Metabolic changes:} 
The lower gravity of Mars, its environment and the nutrition changes will have a big impact on future crews’ metabolisms. Here, a protocol is developed for the monitoring of essential parameters of the health and metabolism of the crew members. 
% This project is lead by \textit{Jean Jacobs} (executive officer).
\textbf{Insects in the astronauts’ diet:} Insects constitute a potential alternative food solution for astronaut crews. The viability and yield rate of three insect species (orthoptera, beetle and lepidopteran) are experimented under Martian conditions.
% This project is lead by \textit{Sirga Drouet} (crew journalist).
\textbf{Human flora bacteria on Mars:} 
The survival of some human flora bacteria and the efficacy of several antibiotics under Martian environmental conditions is experimentally studied.
% This project is lead by \textit{Audrey Comein} (crew scientist).
\textbf{Biofertilizers in Martian soil substrate:} 
This experiment analyzes how a closed environment like the MDRS station and with a Martian regolith, the caloric intake of astronauts can be filled thanks to biofertilizers in small quantities.
% This project is lead by \textit{Cheyenne Chamart} (greenhab officer).

\begin{figure*}
\centering
\includegraphics[width=\textwidth]{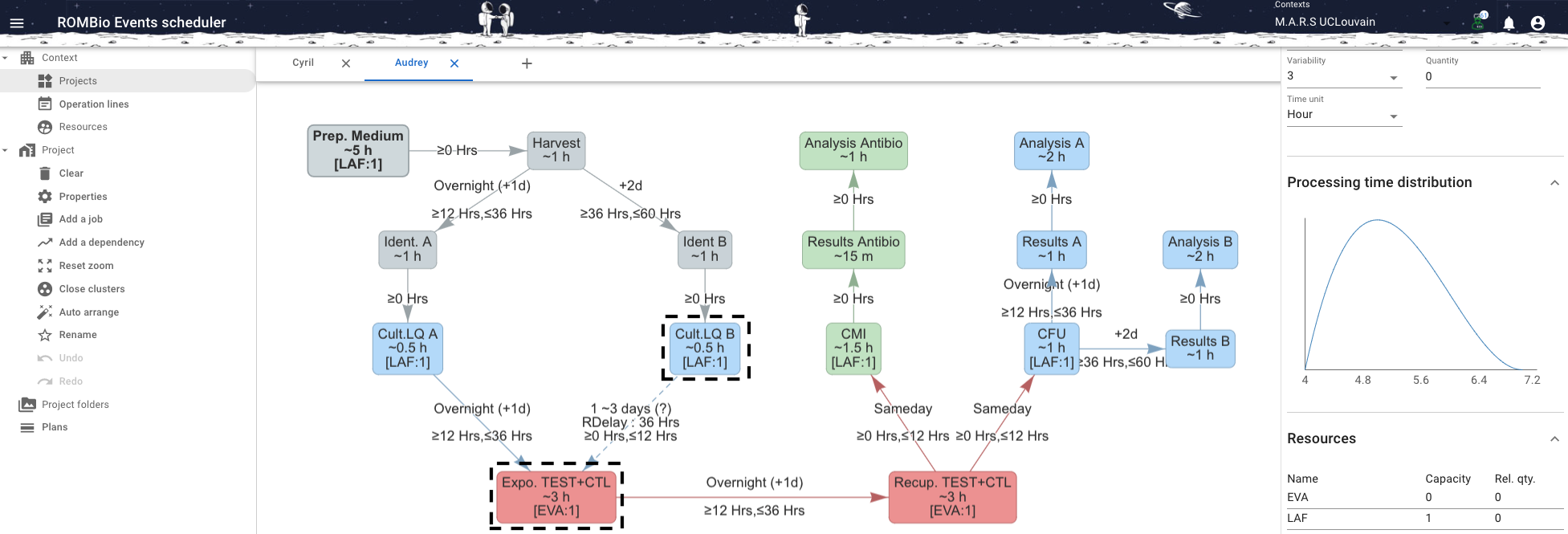}
\caption{Research project \textit{``Survival of human flora bacteria''}. Amongst the activity properties in the right panel, we note that the selected activity \textit{Prep. Medium} requires the \textit{LAF % Spell out the acronym
} resource. The temporal constraint between \textit{Cult.LQ B} and \textit{Expo.TEST+CTL} (dashed) involves a stochastic delay, between 1 and 3 Martian days. }
\label{fig:model_audrey}
\end{figure*}

\subsubsection{Modelling and Scheduling on the RAMS system}

Figure \ref{fig:model_cyril} shows the modelling of one research project, as encoded in \textit{Romie}. 
In fact, this modelling started with a discussion with the experimenter, which lead to a hand-drawn sketch. From this, a first encoding could be made on the system, using the graphical modelling interface, which formally encodes all the activities and constraints involved.

Figure \ref{fig:model_audrey} shows another research project. 
From an operational point of view, this model has interesting properties. It involves a resource shared with other scientists: the laminar air flow (LAF). Since there is only one LAF in the station's science dome, this prevents other activities (belonging to other projects), also requiring the LAF, to be carried out at the same time.
Another point of interest is the temporal constraint between \textit{Cult.LQ B} and \textit{Expo.TEST+CTL}, which involves a stochastic delay. In fact, the delay that must be waited between those two activities (1, 2 or 3 days) depends on the time needed by the bacteria to grow, and it is totally unpredictable by nature. 
Finally, there are temporal constraints, stating that some activity should not start sooner and/or later than a defined amount of time after some other activity. 
In addition to precedence constraints, the temporal constraints present in this model may potentially lead to a project failure, due to the underlying temporal uncertainty.

\begin{figure*}
\centering
\includegraphics[width=\textwidth]{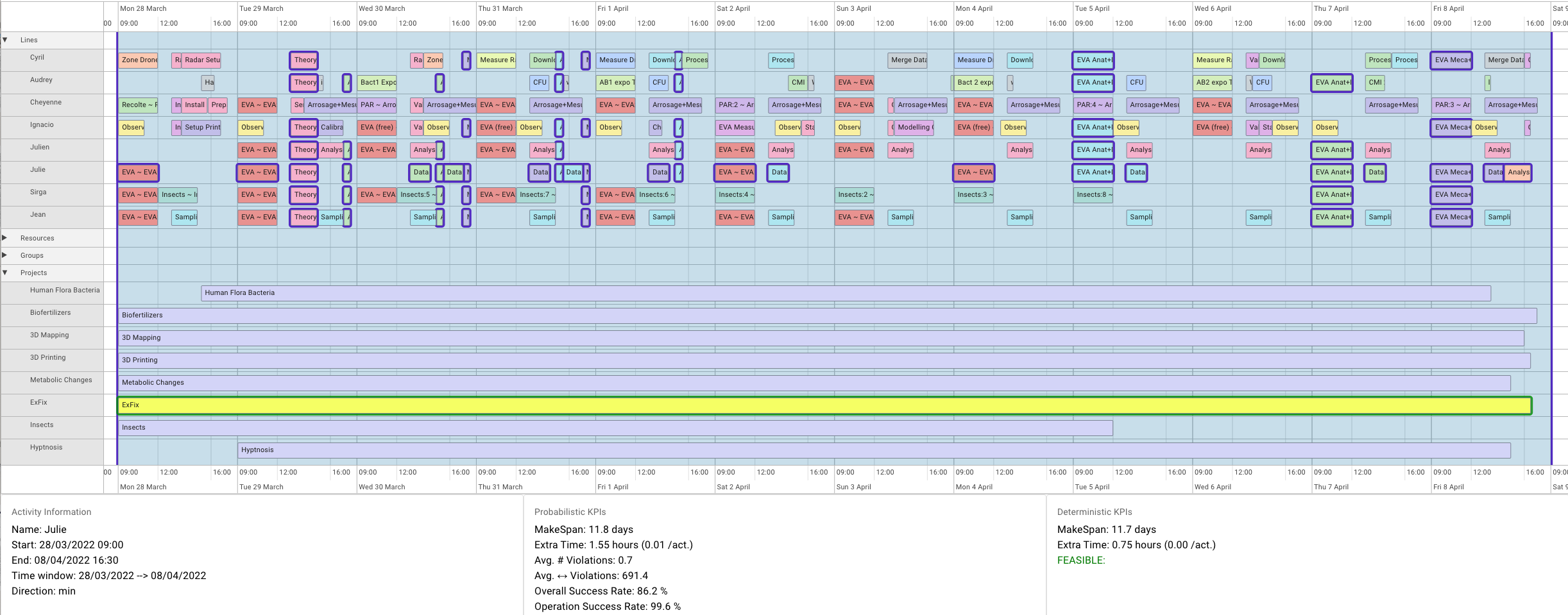}
\caption{A provisional schedule. This is the a priori schedule computed before the beginning of the operations. It involves 162 activities, each having contingent durations. The overall success probability is of 86.2\%. Highlighted in yellow, the \textit{ExFix} research project with all the related activities above, framed with blue rectangles. We see that this project imposes activities to several of (in fact, almost all) the crew members.}
\label{fig:planning_provisional_all}
\end{figure*}

The temporal constraints present in this model may potentially lead to a project failure, due to the underlying temporal uncertainty. 
Yet, another kind of complexity lies in models that involve the participation of several crew members, in addition to shared equipment. 
Figure \ref{fig:planning_provisional_all} shows an example of an optimized \textit{provisional} schedule, as obtained using \textit{Romie}'s optimization engine, for all eight research project during the entire mission. 
In this schedule, the activities involved in research project \textit{ExFix} are highlighted. 
We directly see that many of these require time within the schedule of the other crew members. 

The RAMS system here not only allows to check the deterministic KPIs, but also some probabilistic ones. 
From a \textit{deterministic} point of view, when all the durations are assumed to require their nominal operational time, this planning is \textit{feasible}. For example, in the project modelled in Figure \ref{fig:model_audrey}, provided that the delay between \textit{Cult.LQ B} and \textit{Expo.TEST+CTL} will reveal to be exactly two sols. However, when taking \textit{uncertainty} into account, then the mission probability of success is of 86.2\%. 
Here the system only takes temporal uncertainty into account, not the fact that the activities themselves could be failed, requiring a rescheduling. Rescheduling operations, as well as adding new operations on the fly, will be part of the astronauts' daily manipulation on the system.

\section{Theoretical Foundations} \label{sec:theory}

In stochastic contexts such as space missions, computing optimal schedules becomes significantly less attractive as problem data, such as the manipulation time of the modelled activities, are different from their predicted nominal values. This is what we refer to as \textit{uncertainty}. 
In a constrained environment with shared resources and devices, when they arise such temporal deviations can propagate to the remaining operations, eventually leading to global infeasibility, that is, a project failure.
Given a schedule, a central question is then the following: considering temporal uncertainty, what is the actual probability of success of the mission?

\subsection{Project management is hard}

The problem of scheduling a set of operations under constraints should be seen as a generalization of the well-known \textit{NP}-complete \emph{job-shop scheduling problem} \cite{lenstra1979computational}, which has the reputation of being one of the most computationally demanding (\citeauthor{applegate1991computational}, \citeyear{applegate1991computational}). 
When taking uncertainty into account, the problem then becomes strongly \textit{NP}-hard, an even harder family of problems. 
In a nutshell, \textit{NP}-complete means that, no matter the available computational resources, the problem is conjectured as impossible to solve in practice, for realistic instance sizes, such as the number of activities and resources. 
In fact, whereas the problem depicted in Figure \ref{fig:example} admits only two solutions, in practice the number of possible schedules grows exponentially with the number of tasks and resources. 
Back to Voyager 2 space probe mission, suppose we are interested in all the possible permutations between its 175 operations, then we have $175! \approx 10^{318}$ possible permutations.

\paragraph{Solution methods for combinatorial optimization problems}
Fortunately, algorithmic and mathematical techniques exist in order to solve the problem without enumerating all the $10^{318}$ permutations and schedules. 
Mathematical methods, such as \textit{integer programming}, \textit{constraint programming}, \textit{SAT solving}, and so on are known to be powerful methods, in the sense that given the right formulation of the the problem, generic solvers (e.g. Gurobi, CPLex) are able to find solutions and eventually provide proofs of optimality (without performing an exhaustive enumeration). Unfortunately, finding the correct formulation is usually the most complicated part of the problem, and the languages accepted by these generic solvers do not allow to express complicated specific operational constraints or objectives. 

On the other hand, heuristic methods, such as \textit{local search} which is exploited here in the \textit{Romie} system, or other methods such as \textit{genetic algorithms}, trade the completeness of the exact methods for more flexibility. Such methods will be able to find  (hopefully) good solutions but, even when a solution found by the algorithm is optimal, it will not be able to prove (or even determine) it. However, describing what makes a solution acceptable or not, and what is the quality of a solution, is made much easier when there is no mathematical proof generation framework in behind.

The optimization engine of our decision system \textit{Romie} is based on local search. In a local search algorithm, the key ideas are the following. (S) Start from an initial (potentially infeasible) solution $x$, such as a random permutation. This is the current solution. (M) Apply a local modification to $x$ leading to another solution $x'$, for example, by permuting two activities at random. Then, (E) evaluate the quality $f(x')$ and decide, according to $f(x)$, whether or not $x'$ becomes the new current solution. Finally, repeat (M) and (E) until some stopping criterion (time or solution quality) is met. In the end, return the best solution encountered.

\subsection{Uncertainty management}

As shown in our introductory example, provided two different schedules, determining the best one (\textit{e.g.} the more reliable) in light of the uncertainty is not trivial. 
Now, suppose this must be done for each and every permutation that is considered. 

Computing the probability of success of a given permutation of activities can be done by computing the degree of dynamic controllability, or \textit{robustness}, of the associated \textit{probabilistic simple temporal network} (PSTN). 
Different approaches have been proposed to either approximate or compute this robustness. 
It is important to state that the robustness of a system depends on the uncertainty of course, but also on how clever is the system at reacting to random events. 
In \cite{SaintGuillain2021}, we refer to this ``cleverness'' as the \textit{execution (or dispatching) protocol} $\mathcal{P}$, sometimes called \textit{policy}, which defines how the system reacts to random events. The protocol may consist of simple rules, such as ``start every operation as soon as possible''. It could also be more elaborated strategies, involving preventive waiting times. A particular case is that of a protocol $\mathcal{P}$ solving the perfect online reoptimization problem (a multistage stochastic program). Depending on $\mathcal{P}$, the \textit{robustness} of a system, namely a PSTN, can be defined in general terms as:
\begin{eqnarray}
    r^\mathcal{P}(N) = \sum_{\xi \in \Omega^N} \mathbb{P}\{ \xi \} ~ \Phi^\mathcal{P}(N, \xi) \label{eq:dp_robustness}
\end{eqnarray}
where $\Omega^N$ is the support of all possible scenarios in which the PSTN could fall due to random events, $\mathbb{P}\{ \xi \}$ is the probability to fall into a particular scenario $\xi$, and $\Phi^\mathcal{P}(N, \xi)$ simple returns one if executing the protocol $\mathcal{P}$ in scenario $\xi$ leads to a successful mission, zero otherwise. 
In our case, a scenario $\xi$ is an assignment of a duration to each activity, with support $\{\xi \in \mathcal{R}^n: \mathbb{P}\{ \xi \} > 0\}$.

The computation of (\ref{eq:dp_robustness}) remains intractable in practice, as the size of $\Omega^N$ grows exponentially with the number of contingent operations. For instance, consider the 162 operations involved in the provisional schedule of our M.A.R.S. UCLouvain 2022 mission, as depicted in Figure \ref{fig:planning_provisional_all}. Suppose each operation can take any duration between 20 and 60 minutes, precise to the minute, then we have $20^{40} \approx 10^{88}$ scenarios possible.
This motivates all kinds of sampling based methods, such as Monte Carlo, which therefore restricts the summation in (\ref{eq:dp_robustness}) to a limited subset of $S \subset \Omega^N$. 

In \cite{SaintGuillain2021}, we showed that by reasoning on the connections between the random variables, instead of the scenarios, then the exact robustness could be computed exactly, in pseudo-polynomial (\textit{i.e.} efficient) time. However, this is only provided that $\mathcal{P}$ actually is the natural ``as soon as possible'' strategy (and nothing more elaborated), and that the PSTN is well formed. 
The PSTN formalism being quite restrictive, there are many considerations of a space mission that cannot be modelled as a well formed PSTN, such as exotic constraints and resource issues (e.g. energy consumption). 
On the contrary, sampling-based methods are much more versatile, they generally do not impose a formalism as strict as PSTN's. 
Our \textit{Romie} decision system implements such a method, the sample average approximation (\citeauthor{kleywegt2002sample}, \citeyear{kleywegt2002sample}), within its optimization engine.

\subsection{System architecture}
\begin{figure}
\centering
\includegraphics[width=.8\columnwidth]{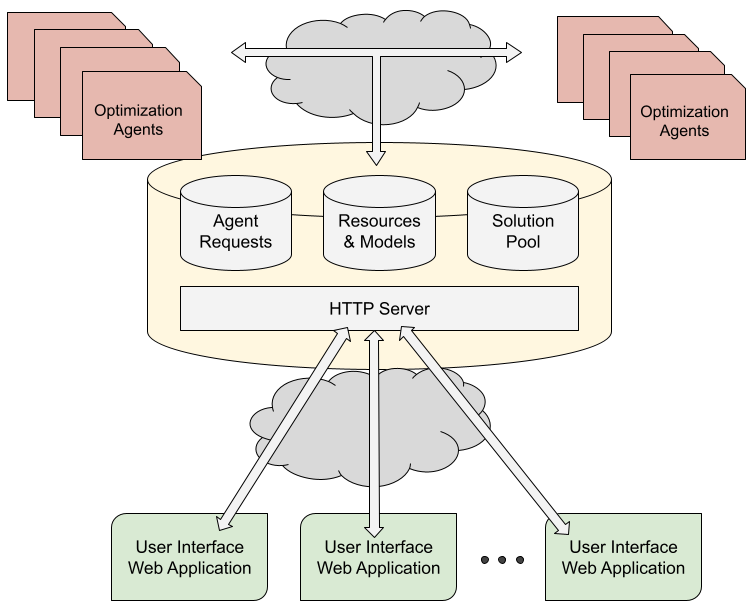}
\caption{Global architecture of the Romie robust advanced modelling and scheduling (RAMS) system.}
\label{fig:architecture}
\end{figure}

Romie uses a decentralized architecture, as depicted in Figure \ref{fig:architecture}.
The user interface is decoupled from the optimization engine part. 
There can be potentially many users connected to the system (\textit{e.g.} the whole team of astronauts), using a classical web browser, since the user interface is implemented using a recent web technology (React).
The optimization engine is actually composed of an arbitrary number of optimization agents, not necessarily hosted on the same servers, hence enabling parallel computing.

The whole system is organized upon a database, which enables the end users to communicate with the optimization agents, but also the optimization agents to communicate altogether in order to distribute and parallelize the work. 
The database stores the following three key information: 
\begin{itemize}
    \item \textbf{Resources \& Models}. Stores the operational context and problem descriptions, namely the models (\textit{e.g.} the scientific project modelled in Figure \ref{fig:model_audrey}) describing the operational projects at stake. The user interface provides a visual representation of the logical information stored in the database, which is rather described in terms of discrete mathematical structures such as sets and graphs. 
    \item \textbf{Solution pool}. The current states of computations. The solution pool is used by the optimization agents to communicate and share their results. An element of the pool is naturally a schedule (e.g. Figure \ref{fig:planning_provisional_all}). For the same optimization problem, the pool may contain several (\textit{e.g.} 10) solutions, corresponding to the bests solutions found so far by all the agents. 
    \item \textbf{Agent requests}. A list of requested user actions. An action could be either one-shot, such as adding or removing a project for instance, or a running action, such as optimizing. One-shot requests are picked up by exactly one idle agent (at random), which will perform the associated action (\textit{e.g.} remove a specific operation from a given project in a given schedule, and then recompute the KPIs, such as the overall probability of success). The only possible running request is an optimization action, which triggers all the available idle agents to concurrently try to improve a given schedule. 
\end{itemize}
Whereas the architecture is designed to be decentralized, in practice all the elements could easily be integrated on the same computer, or machine (such as a spacecraft). 

The communication formalism between the computing agents is however robust to latency, which enables a same physical system to host its own limited set of agents, whereas remote agents can be solicited in support, even with important delays (\textit{e.g.} a couple of seconds from Earth to Moon). Or a fleet of autonomous planetary rovers could distribute the computational effort on any possible computational support in an acceptable range (say, five to ten light seconds), including the rovers themselves. 

Also, the system easily recovers (actually, is not impacted at all) from the disconnection and the death of optimization agents, and new agents can be added on the course of the computations.
Human operators connect remotely to the system, using a simple web browser, and may disconnect and reconnect without loss of any information, and without interrupting any ongoing computation. Finally, several different computations, for instance optimizing the same initial planning under three different combinations of KPI preferences, or under different operational resource limitations, may be carried out simultaneously (in which case these different optimization problems are distribution over all the available agents).

\subsection{Online reoptimization} \label{sec:reoptimization}

\begin{figure*}
\centering
\includegraphics[width=\textwidth]{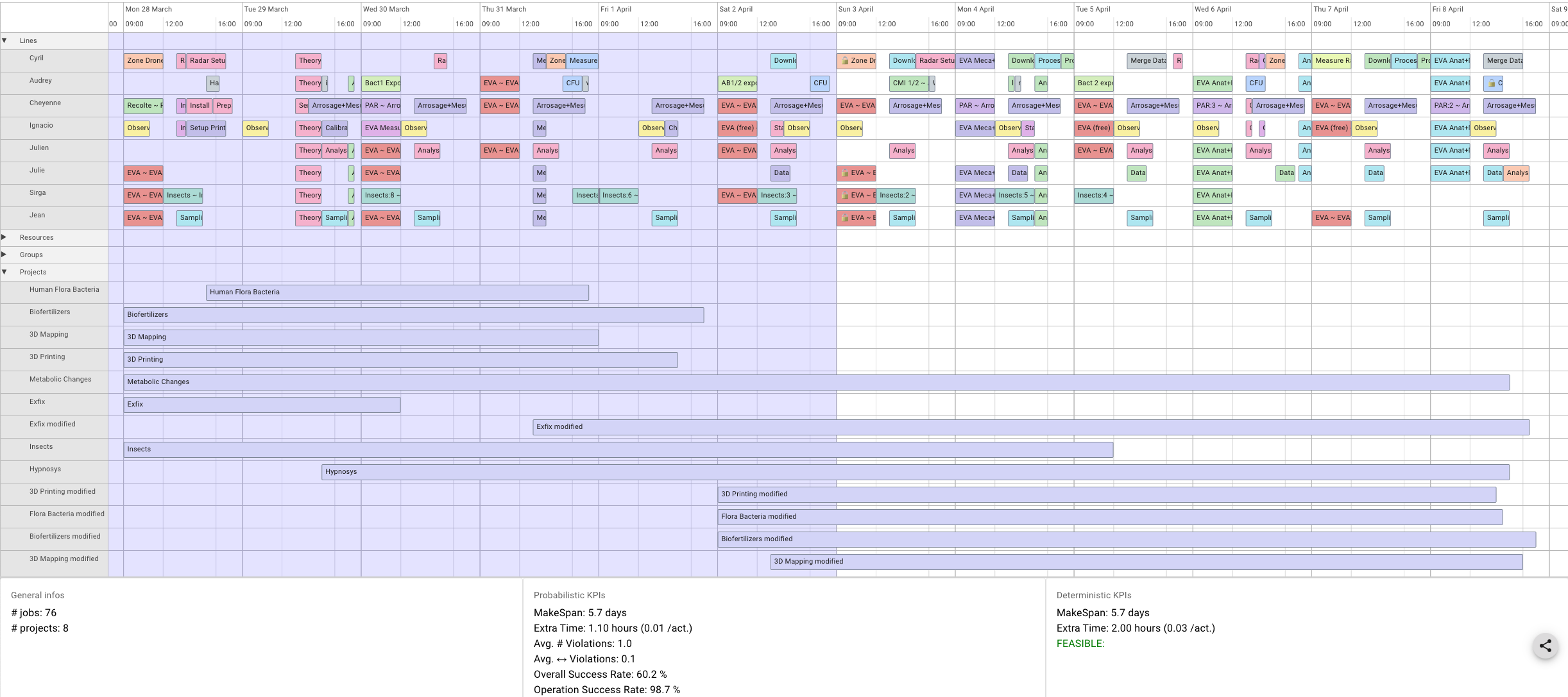}
\caption{Current schedule at Sol 6. Many scientific projects had to be remodeled. As the operations are conducted, the project models are likely to be adapted by the astronauts as some constraints or activities reveal to be inadequate to actual real world conditions. For example, the ExFix project has been interrupted on March 30th, and rescheduled from then on, based on a modified model. }
\label{fig:planning_sol6}
\end{figure*}

Whereas the provisional schedule depicted in Figure \ref{fig:planning_provisional_all} is computed before the beginning of the operations, in practice things rarely happen exactly as initially planned.
In comparison, Figure \ref{fig:planning_sol6} shows the current state of past (executed) and future (planned) operations, at Sol 6. 
Remark for instance that most of the activities initially planned at Sol 2, during the morning, disappeared and had to be rescheduled. In fact, these corresponded to an EVA that had to be cancelled, due to bad weather conditions.

This leads to an online (\textit{i.e.} dynamic) re-optimization problem, in which the decisions must be optimized given a fixed current state of the system, including past activities. During the mission, the astronauts actually updated their schedule at the end of every day, encoding what really happened, and reoptimized for the rest of the mission. 

The best one can do is therefore to compute the schedule that is the more likely to succeed, knowing that in fact, no one knows what will actually happen. Mathematically speaking, this can be described as playing a game against Nature, in which at current time $t$, we take the decisions $x^t$ that maximize their expected outcome (\citeauthor{saint2021lila}, \citeyear{saint2021lila}), leading to the following multistage stochastic program:
\begin{multline}
    \underset{x^t \in X^{t}}{\text{argmax}} ~~ E_{\xi^{t+1}} \Big[ \max_{x^{t+1} \in X^{t+1}} E_{\xi^{t+2}} \\ 
    \Big[  \ldots \max_{x^{h-1} \in X^{h-1}} E_{\xi^{h}} \Big[\max_{x^{h}  \in X^{h}} V^{x^{1..h}}(N, \xi) \Big] \ldots \Big] \Big]
        \label{eq:multistage}
\end{multline} 
where the maximum value of the first expectation is, by definition, equal to the current probability of success under perfect reoptimization. 

\begin{figure}[t]
\centering
\includegraphics[width=\columnwidth]{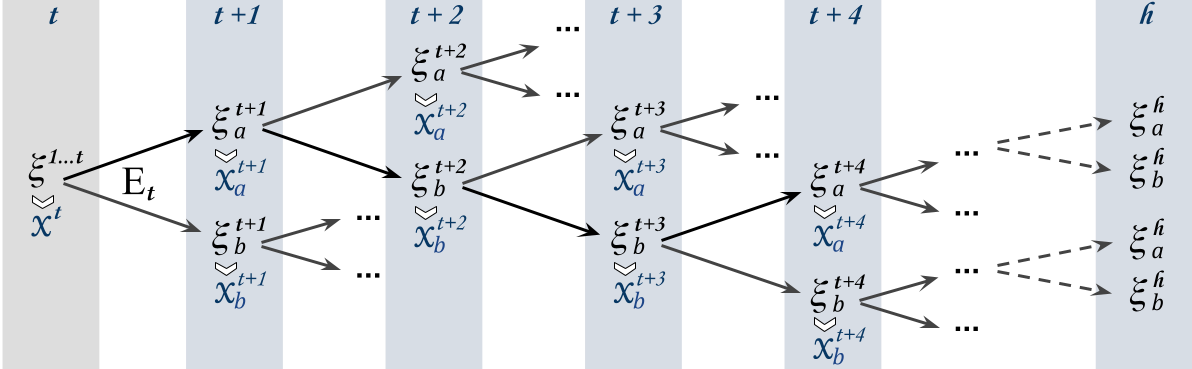}
\caption{Tree structure of the problem. The root node represents the current state (past decisions and realizations) at time $t$. For simplicity, decision (\textit{resp.} random) variables have only two possible choices (\textit{resp.} outcomes).}
\label{fig:scenarios}
\end{figure}

The nested expectations in (\ref{eq:multistage}) form a tree structure, well known as the \textit{scenario tree}. Unfolding the maximization operators as well leads to a full decision-scenario tree as illustrated in Figure \ref{fig:scenarios}. 
Each path of the tree constitutes a possible scenario realization together with associated decisions, a sequence $\xi^t, x^t, \ldots, \xi^h, t^h$. At time $t$, decisions $x^t$ depend on the current history $\xi^{1..t}$ and maximize the expected value $E_{\xi^{t+1}}[\max_{x^{t+1}} \ldots]$ of the future decisions at time $t+1$ given the remaining uncertainty, and so on until time $h$ is reached. 

The number of scenario  $\xi \in \Omega$ in equation (\ref{eq:dp_robustness}) is equal to the number of different paths in Figure \ref{fig:scenarios}. 
We directly see that the size of our scenario tree grows exponentially with the number of decision steps and outcomes, which explains why computing (\ref{eq:dp_robustness}) is intractable in practice.

\subsection{Previous researches} \label{sec:previous_researches}

Based on the real case study of a Mars analogue mission in 2018, in \cite{SaintGuillain2019} we proposed a first (incomplete) probabilistic formulation, as well as solution method, for the problem of scheduling a set of various human operated projects.
In fact, the problem of scheduling a set of operations in a constrained context such as the \emph{Mars Desert Research Station} (MDRS, Fig. \ref{fig:mdrs}) is not trivial, even in its classical deterministic version. 
We hence measured the gains and costs, on a priori mission planning, of robust schedules (optimized under uncertainty) compared to schedules optimized under classical deterministic assumptions.

In \cite{SaintGuillain2021}, the theoretical insights obtained from the former study were successfully extended to \textit{probabilistic simple temporal networks} (PSTNs), a formalism able to mathematically describe operational problems in general, such as scheduling a space mission or a biomanufacturing campaign. 
In this paper written with the Jet Propulsion Lab (NASA), our probabilistic model is applied to the operation management of Mars 2020 planetary rover. We also formally define some of the most important theoretical concepts for describing schedule robustness to uncertainty, we introduce new ones, and give proofs for theoretical bounds.
This contributed to filling the theoretical gap between specific mission planning and general operations management. 

Finally, an asymptotically optimal approach to robustness computation and online reoptimization is briefly explored in \cite{saint2021lila}. In the later study, the problem of dynamically dispatching the activity execution times is modelled as a \textit{single-player game against Nature}, and solved using Monte-Carlo tree search (MCTS). This only constitutes a preliminary study, which still has to be further studied. 

\paragraph{On novice self-scheduling}
These previous studies mainly aimed at evaluating and demonstrating, both empirically and theoretically, the need and the advantages of using probabilistic assumptions (\textit{i.e.} optimizing under uncertainty) in the context of operations management. 
In the context of space exploration, past missions (\textit{e.g.} UCL to Mars 2018, \citeauthor{SaintGuillain2019} \citeyear{SaintGuillain2019}) have shown the importance of online reoptimization and, in particular, the need for the crew to autonomously adapt their science projects to unforeseen events. 
In the current paper, the scientific focus is rather put on the user experience. More specifically, using techniques from human computer interaction (HCI), we measure how well a team of novice users succeeded (or not) at using our system to schedule, and reschedule online, their own activities. 

% \begin{figure}
% \centering
% \includegraphics[width=.9\columnwidth]{figures/MDRS_EVA.png}
% \caption{Left: the Mars Desert Research Station in Utah. Right: extra-vehicular field operations.}
% \label{fig:MDRS}
% \end{figure}

\subsection{A new risk-aversion paradigm}
\textit{What if analysis} and \textit{sensitivity analysis} are classical, well-known techniques for coping with uncertainty in operations management. 
In fact, \cite{papavasileiou2007optimize} and \cite{petrides2014biopharmaceutical} both argue for the importance of simulation and the ability of performing what-if and/or sensitivity analysis in addition to optimization.
What if analysis consists of optimizing several solutions (usually a few numbers), each solving a predefined scenario, such as best-case, average-case and worst-case scenarios. 
In \cite{SaintGuillain2021} we formally prove that the well known \textit{what-if analysis} technique is fundamentally flawed, as it arbitrarily underestimates a schedule's risk.
The degree of weak controllability (DWC) can be interpreted as a perfect what-if analysis, that is, when not only considering best, middle and worst cases, but all the possible scenarios. In the later study, the demonstration is then reached at inequality (11), with the result $\text{DDC}(N) \le \text{DWC}(N)$, where DDC (degree of dynamic controllability) is the true maximal robustness of the schedule $N$.
On the contrary, Romie’s optimization engine is proven to never underestimate the risk. 

Provided a schedule (this however does not help at finding the right schedule in the first place!), \textit{sensitivity analysis} approximates its average quality under uncertainty, how sensible (brittle) it is to stochastic variations. 
In that sens, the solutions computed by Romie directly optimize their response to a sensitivity analysis.
The proposed RAMS framework introduces a new paradigm, replacing both what-if and sensitivity analysis.

\section{Astronaut's ability to self-schedule}
\label{sec:self}

We finally dive into our actual research question: how well the astronauts succeed at self-scheduling their scientific operations, provided a RAMS decision-support system such as \textit{Romie}. 
The astronauts were asked to evaluate their experience of the system and the quality of the computed decisions, before and at different stages during the mission. 
The a priori stage, before the mission, is called \textit{sol zero} ($S_0$). A \textit{sol} is a day on Mars.
The subsequent stages are $S_4, S_8$ and $S_{12}$, for sols four (early mission), eight and twelve (end of the mission).

We collected demographic information from the participants at sol zero ($S_{0}$), that is \textit{before the beginning of the mission}. Participants were instructed to complete an UEQ+ questionnaire (User Experience Questionnaire)~\citep{Schrepp:2019}, a modular extension of the UEQ evaluation method in which we selected 12 scales (i.e., \textsc{Attractiveness}, \textsc{Efficiency}, \textsc{Perspicuity}, \textsc{Dependability}, \textsc{Stimulation}, \textsc{Novelty}, \textsc{Trust}, \textsc{Adaptability}, \textsc{Usefulness}, \textsc{Visual aesthetics}, \textsc{Intuitive use}, and \textsc{Trustworthiness of content}) among 20 scales to focus on evaluating the user experience of participants interacting with the system.
Each scale is in turn decomposed into four subscales or items to be evaluated (e.g., attractiveness is decomposed into four subscales: annoying vs. enjoyable, bad vs. good, unpleasant vs. pleasant, and unfriendly vs. friendly), each subscale being a differential scale with 7 points between items of each pair (e.g., annoying $\rightarrow$ enjoyable). We measure each item employing a 7-point Likert-type scale with response categories ``Strongly disagree'' (=1) to ``Strongly agree'' (=7).

\begin{figure*}
\centering
% \vspace{-8pt}
\includegraphics[width=\textwidth]{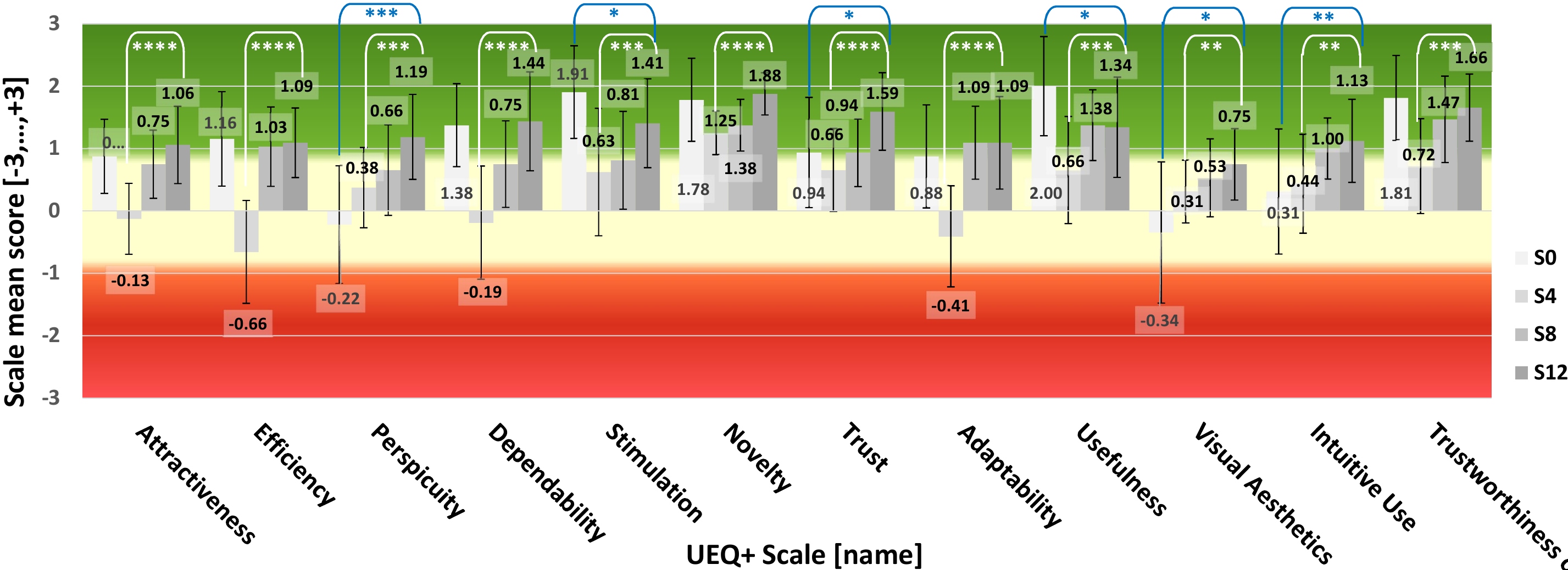}
\vspace{-18pt}
\caption{UEQ+ scale mean scores for all sessions $S_{i}$. Significant differences between between $S_0$ and $S_{12}$ are represented in blue on the top, between $S_4$ and $S_{12}$ are represented in white. Error bars show a confidence interval of 95\%.}
\label{fig:scale_means_all}
\end{figure*}

UEQ+ was selected as an evaluation method because it is a modular and modern interface evaluation method where scales can be decided based on the interface to evaluate and covers the user experience (UX), not just usability, as assessed by questionnaires such as IBM PSSUQ~\citep{Lewis:2006}. UEQ+ is also easy to administer to participants and remains valid even with a limited number of participants. For some scales, a benchmarking of their values leads to an interpretation of five effect sizes~\citep{Schrepp:2017}: bad, below average, above average, good, and excellent.  
Consequently, for each sol $S_0, S_4, S_8, S_{12}$, we have two additional dependent variables:
\begin{enumerate}
    \item The \textsc{Scale mean score}, a real variable measuring the average score obtained on all items of each scale.
    \item The \textsc{Scale mean importance}, a real variable measuring the average weight of importance of each scale.
\end{enumerate}
Participant answers are interpreted with the \href{https://ueqplus.ueq-research.org/Material/UEQ_Plus_Data_Analysis_Tool.xlsx}{UEQ data analysis tool}. 
According to \citep{Schrepp:2019}, ``it is extremely unlikely to observe values above +2 or below -2,..., the standard interpretation of the scale means is that values between -0.8 and 0.8 represent a neutral evaluation of the corresponding scale, values superior to 0.8 represent a positive evaluation, and values inferior to -0.08 represent a negative evaluation``. The same interpretation holds for the scale mean importance.
Fig.~\ref{fig:scale_means_all} shows the scale means for all sessions $S_0, S_4, S_8, S_{12}$ with their corresponding mean importances. 

\subsection{Before the mission: $S_0$}

Considering the measurements done at sol zero ($S_{0}$), that is \textit{before the beginning of the mission}, only two scales of twelve are negatively assessed in the neutral zone. 
First, \textsc{Perspicuity} ($M{=}{-}0.22$, \textit{SD}${=}1.36$) expresses that the participants did not quickly familiarize themselves with the system, which they nevertheless judged to be very important ($M{=}1.88$, \textit{SD}${=}0.60$), since this was their first discovery of the system.
Second, \textsc{Visual aesthetics} ($M{=}{-}0.34$, \textit{SD}${=}1.63$) was also negatively assessed for user interface aspects estimated unimportant with the lowest score ($M{=}{-}0.38$, \textit{SD}${=}1.58$). \textsc{Intuitive use} ($M{=}0.31$, \textit{SD}${=}1.45$) is the only scale assessed positively in the neutral zone although very important too ($M{=}2.13$, \textit{SD}${=}0.60$). However, three scales were borderline, that is, \textsc{Attractiveness} ($M{=}0.86$, \textit{SD}${=}0.88$), \textsc{Adaptability} ($M{=}0.88$, \textit{SD}${=}1.19$), and \textsc{Trust} ($M{=}0.94$, \textit{SD}${=}1.27$), thus reflecting that participants were still not convinced that the system fulfilled their needs for these three important aspects. Fortunately, six of 12 scales are positively assessed, even above the threshold, thus suggesting that the participants felt these aspects are already well fulfilled at first glance:
\textsc{Usefulness} ($M{=}2.00$),
\textsc{Stimulation} ($M{=}1.91$),
\textsc{Trustworthiness} ($M{=}1.81$),
\textsc{Novelty} ($M{=}1.78$),
\textsc{Dependability} ($M{=}1.38$), and
\textsc{Efficiency} ($M{=}1.16$).

\subsection{During the mission: $S_4$, $S_8$, $S_{12}$}
We now consider the scales for $\boldsymbol{S_4}$, that is, at an early stage of the mission, at the end of the fourth day.
Only \textsc{Novelty} ($M{=}1.25$) exceeds the 0.8 threshold with small dispersion (\textit{SD}${=}0.50$), thereby meaning that participants recognise that the software was original, partly because they were never confronted to any similar software. Seven of 12 scales are positively assessed in the neutral interval, representing a slight improvement with respect to $S_0$:
\textsc{Trustworthiness of content}  ($M{=}0.72$, $SD{=}1.10$)
\textsc{Trust} ($M{=}0.66$),  
\textsc{Usefulness} ($M{=}0.66$),
\textsc{Stimulation} ($M{=}0.63$),
\textsc{Intuitive Use} ($M{=}0.44$),
\textsc{Perspicuity} ($M{=}0.38$), and
\textsc{Visual Aesthetics} ($M{=}0.31$).
The three most positive scales refer to the utility character of the application, which is considered as the most important part.
Four scales are negatively assessed in the neutral interval, thus calling for improvement:
\textsc{Efficiency} ($M{=}{-}0.66$),
\textsc{Adaptability} ($M{=}{-}0.41$),
\textsc{Dependability} ($M{=}{-}0.19$), and
\textsc{Attractiveness} ($M{=}{-}0.13$).

While \textsc{Novelty} received the highest mean score, it also received the lowest importance ($M{=}{-}0.13$), because participants become more accustomed with the software and therefore reduce its importance over time. Similarly, \textsc{Visual aesthetics} ($M{=}0.25$) were no longer considered as important as before. Utility scales take precedence over usability scales in terms of mean importance:
\textsc{Efficiency} ($M{=}2.38$),
\textsc{Usefulness} ($M{=}2.25$),
\textsc{Perspicuity} ($M{=}2.13$),
\textsc{Trustworthiness of content}  ($M{=}1.88$),
\textsc{Adaptability} ($M{=}1.88$),
\textsc{Dependability} ($M{=}1.75$),
\textsc{Intuitive Use} ($M{=}1.75$),
\textsc{Trust} ($M{=}1.50$), 
\textsc{Attractiveness} ($M{=}1.13$), and
\textsc{Stimulation} ($M{=}0.88$).

\begin{figure*}
\centering
%    \vspace{-8pt}
\includegraphics[width=\textwidth]{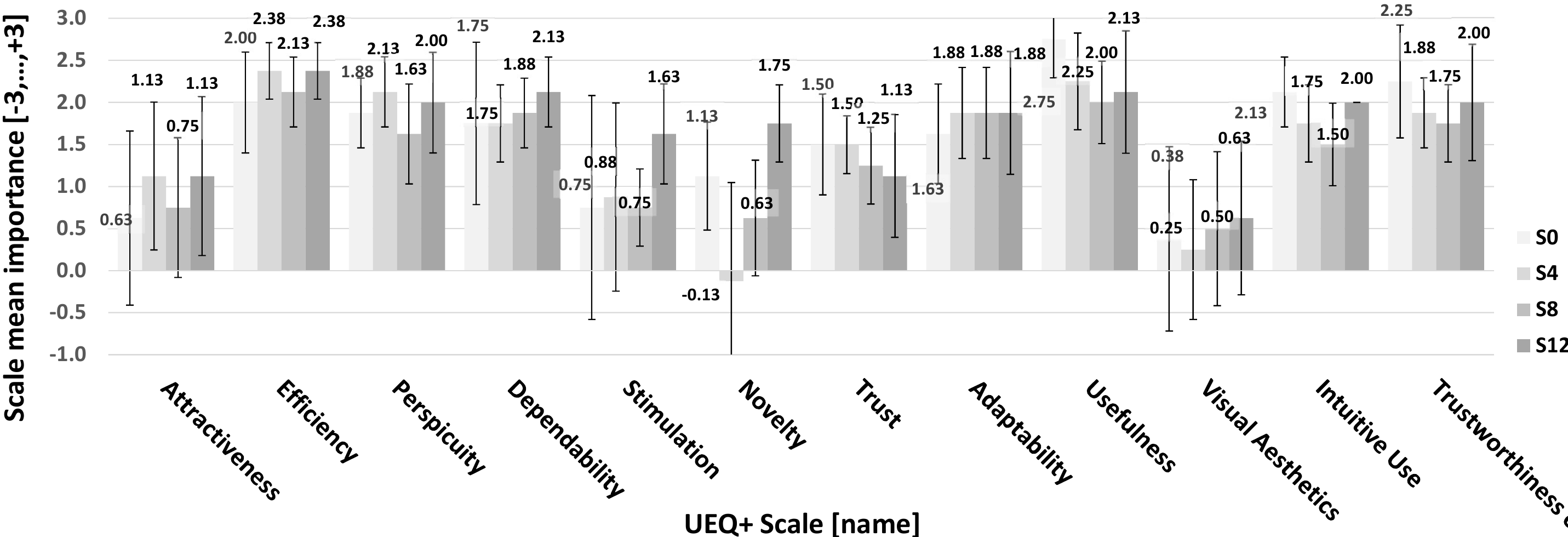}
\vspace{-18pt}
\caption{UEQ+ mean importance for all sessions $S_{i}$. Error bars show a confidence interval of 95\%.}
\vspace{-8pt}
\label{fig:importance_ratings_all}
\end{figure*}

For the first time, at sol eight $\boldsymbol{S_8}$, all scales become positively assessed with only four belonging to the neutral zone: 
\textsc{Attractiveness} ($M{=}0.75$),
\textsc{Dependability} ($M{=}0.75$),
\textsc{Perspicuity} ($M{=}0.66$), and
\textsc{Visual aesthetics} ($M{=}0.53$).
The remaining eight scales are located above the threshold: 
\textsc{Trustworthiness of content} ($M{=}1.47$),
\textsc{Usefulness} ($M{=}1.38$),
\textsc{Adaptability} ($M{=}1.09$),
\textsc{Novelty} ($M{=}1.38$),
\textsc{Efficiency} ($M{=}1.03$),
\textsc{Intuitive use} ($M{=}1.00$),
\textsc{Trust} ($M{=}0.94$), and
\textsc{Stimulation} ($M{=}0.81$). \textsc{Trustworthiness} received the highest scale means and a high importance ($M{=}1.75$), thereby suggesting that participants progressively acquire more trust in manipulating the data. The functions attached to these data are well perceived based on \textsc{Usefulness} with a high importance ($M{=}2.00$). \textsc{Efficiency} ($M{=}2.13$, $SD{=}0.60$) was rated the most important factor although its scale was not the highest one.

The last session $\boldsymbol{S_{12}}$ (sol twelve, last day of the mission) obtained all scale means above the threshold, thereby indicating the most positive appreciation of the software, except for \textsc{Visual Aesthetics} ($M{=}0.75$, \textit{SD}${=}0.83$), which is also consistently rated as the least important factor ($M{=}0.63$, \textit{SD}${=}1.32$). Surprisingly, \textsc{Novelty} obtained the highest scale mean ($M{=}1.88$) with the smallest deviation (\textit{SD}${=}0.48$), with a moderate importance ($M{=}1.75$, \textit{SD}${=}0.66$), thus suggesting that participants estimate that the software stays original, even after several usages. \textsc{Trustworthiness of content} remains the second highest scale means ($M{=}1.66$, \textit{SD}${=}0.77$) as it was the case before, with a very high importance ($M{=}2.00$, \textit{SD}${=}1.00$). \textsc{Attractiveness} ($M{=}1.06$, \textit{SD}${=}0.90$) suffered from the lowest mean with the second lowest importance rate ($M{=}1.13$, \textit{SD}${=}1.36$), thus suggesting that this factor does not deteriorate much the overall software quality. Just before this factor, 
\textsc{Adaptability} ($M{=}1.09$, \textit{SD}${=}1.07$) and \textsc{Efficiency} ($M{=}1.09$, \textit{SD}${=}0.80$) share the second lowest scale mean, with the highest importance for \textsc{Efficiency} though ($M{=}2.38$, \textit{SD}${=}0.48$).

\subsection{Inter-session evolution: Results and Discussion}
In this section, we first use statistical tools (inter-rater agreement, inter-rater consistency) to address two questions:
Do we have obvious consensus, or did the astronauts answer independently?; 
Did the participants answer in a random fashion, or based on logical assumptions?
We use two statistical tools: Kendall's coefficient of agreement, and Cronbach's coefficient of consistency. 
The later measures how relevant is the measurement tool (form), while the first quantifies the quality of the sample (group of participants).
Finally, we focus on \textit{the evolution of the results}, during the mission, from Sol 0 (before the mission) to Sol 12 (end of the mission). 

\subsubsection{Inter-rater agreement} 
Table~\ref{tbl:reliability} reports Kendall’s coefficient of concordance $W$ (\cite{Legendre:2005}), a measure of agreement among raters which is equal to 0 when there is no agreement among them. 
The lower the concordance, the more heterogeneous is the survey sample (participants). A high coefficient reflects either one (or several) of the following facts: the sample is too small, the participants were selected with a bias, the participants communicated while answering the questions.

Although all $W$ values are positive, some of them are low, indicating that there is limited agreement (e.g., \textsc{Dependability} in $S_0$ received $W{=}0.041$ interpreted as poor agreement and \textsc{Visual aesthetics} in $S_0$ received $W{=}0.21$ interpreted as fair agreement). Some others depart more from 0, but rarely in a significant way. In particular, \textsc{Usefulness} in $S_0$ received $W{=}0.34$ with $p{=}.041^{*}$, which is the scale benefiting the most from inter-rater agreement in a significant way, thus rejecting the null hypothesis that there is no agreement among participants.  $W$  slightly decreases across sessions, but stays interpreted as 'fair.' Another example is \textsc{Efficiency}, which received a fair agreement ($W{=}0.40$, $p{=}.021^{*}$) for $S_4$, but decreases over sessions. The limited agreement can be partially explained by the diversity of the profiles of the participants, but also by the varying experimental conditions: $S_0$ was carried out in a room with limited pressure, while $S_4$, $S_8$, and $S_{12}$ were carried out with more mental, temporal, and physical pressure.

\subsubsection{Inter-rater consistency}
Table~\ref{tbl:consistency} reports Cronbach's $\alpha$ coefficient computed to quantify the internal consistency, which expresses the extent to which the scale measurements remain consistent within a session or over subsequent sessions under identical or different conditions. 
This test measures how the different components of the form permit to well reflect the global user experience, or in contrary, if the questions asked in the form are not relevant. 
A value above 0.7 is considered as acceptable~\citep{nunnally1975psychometric}, meaning that the scale is most likely to be relevant for the study.

Again, $S_0$ was conducted in lab, while $S_4$ to $S_{12}$ were conducted under experimental conditions mimicking the target real conditions. Individually speaking, scales' $\alpha$ range from an unacceptable interpretation (e.g., \textsc{Perspicuity} received $\alpha{=}0.48$ for $S_8$) to an excellent interpretation (e.g., \textsc{Adaptability} received $\alpha{=}0.93$ for $S_{12}$). $S_0$ obtained 8 values above the 0.7 threshold and 4 values below. The first real condition session, i.e. $S_4$, obtained a balanced consistency: 6 values above the threshold and 6 below. This balance evolves positively across sessions in favour of a consistency above the threshold: 7 above and 5 below for $S_8$ to 9 above and 3 below for $S_{12}$, thus suggesting that participants rated scales more consistently over time. Indeed, the mean $\alpha$ starts at 0.79 (acceptable) for $S_0$ and always improves session after session: 0.61 (questionable) for $S_4$, 0.70 (acceptable again) for $S_8$ and 0.80 (good) for the final $S_{12}$.

\subsubsection{Evolution of scales and importance rates \label{sec:results_evolution}}
Fig.~\ref{fig:scale_means_all} and Fig.~\ref{fig:importance_ratings_all} show how the mean scores and importance evolved across all four sessions. Overall, most scales obtained a high mean for the first  $S_0$, which dramatically decreased for $S_4$ carried out in real conditions, revealing a different appreciation of the software between the ideal conditions \textit{in vitro} and the real conditions \textit{in vivo}. Fortunately, these mean scores positively evolved until reaching positive values above the threshold during the last session. These results suggest that participants, although they were probably influenced by the difficult conditions of $S_4$, progressively improved their assessment, being less influenced by these contextual constraints and more accustomed to deal with them. The results obtained for the last sessions $S_{12}$ therefore represent an overall stable assessment of the software after several continuous usages.

More precisely, Table~\ref{tbl:diff_percentage} shows how scale mean scores and their mean importance evolved in terms of difference of percentage between sessions: first, between the initial \textit{in vitro} $S_0$ session and the first \textit{in vivo} $S_4$ session, then between the two next iterations and, finally, between the first session $S_0$ and the last session $S_{12}$. 
Some scales largely improved since the beginning: \textsc{Perspicuity} received the best mean gain from one session to the last ($\Delta{=}643\%$), followed by \textsc{Visual aesthetics} ($\Delta{=}318\%$) and \textsc{Intuitive use} ($\Delta{=}260\%$), suggesting that the user experience gained during the sessions positively impacted these scale means, even if their mean importance changed over time. Four scales decreased between the first and the last session: \textsc{Usefulness} is reduced by $\Delta{=}{-}33\%$, followed by \textsc{Stimulation} by $\Delta{=}{-}26\%$, \textsc{Trustworthiness of content} by $\Delta{=}{-}9\%$, and \textsc{Efficiency} by $\Delta{=}{-}5\%$, suggesting that participants expressed their needs at a higher level of expectation during the first session than during the last one. This does not depreciate the overall user experience of the interface, but indicates that the experience accumulated by participants let them to adjust their assessment more precisely since all scale means in $S_{12}$ were highly positive (Fig.~\ref{fig:scale_means_all}). Participants also increased their importance rates of nine scales and decreased the rates for three scales only: \textsc{Trust} by $\Delta{=}{-}25\%$, \textsc{Usefulness} by $\Delta{=}{-}23\%$, and \textsc{Trustworthiness of content} by $\Delta{=}{-}11\%$, suggesting that participants have lowered the importance due to the experience gained and the rapid learning curve. The progress acquired during successive sessions is therefore a determining factor for the adjustment of the scales and their importance to converge towards an equilibrium representing a stable value after a continuous interaction.
We investigated whether these differences are statistically significant by computing a Wilcoxon signed-rank test for paired samples between $S_0$ and $S_{12}$, then between $S_4$ and $S_{12}$.

\paragraph{Between $S_0$ and $S_{12}$} \textsc{Perspicuity} is significantly lower ($p{=}.00064^{***}$) with a small effect size ($r{=}0.34$), thus suggesting that participants felt that they were in control much more in the end of the mission than before ((Fig.~\ref{fig:scale_means_all}-blue top bars);
\textsc{Stimulation} is significantly smaller ($p{=}.049^{*}$) with a small effect size ($r{=}0.21$),
\textsc{Trust} is significantly smaller ($p{=}.013^{*}$) with a small effect size ($r{=}0.28$),
\textsc{Usefulness} is significantly larger ($p{=}.020^{*}$) with a small effect size ($r{=}0.25$),
\textsc{Visual aesthetics} are significantly smaller ($p{=}.0037^{**}$) with a small effect size ($r{=}0.33$),
\textsc{Intuitive use} is significantly larger ($p{=}.0049^{**}$) with a small effect size ($r{=}0.32$). With respect to importance, no significant difference was found between means of all corresponding scales, thus suggesting that participants estimated the importance of the respective scales not in a very different way.

% \begin{figure*}
% \centering
% %    \vspace{-8pt}
% \includegraphics[width=\textwidth]{figures/Significance.pdf}
% \vspace{-12pt}
% \caption{Significant differences between scales ratings: between $S_0$ and $S_{12}$ (blue), between $S_4$ and $S_{12}$.}
% \label{fig:significance}
% \end{figure*}

\paragraph{Between $S_4$ and $S_{12}$} Many scales saw their mean scores significantly higher in $S_{12}$ than in $S_4$ (Fig.~\ref{fig:scale_means_all}-white):
\textsc{Attractiveness} is significantly different ($p{\leq}.0001^{****}$) with a medium effect size ($r{=}0.51$),
\textsc{Efficiency} is different ($p{\leq}.0001^{****}$) with a medium effect size ($r{=}0.54$),
\textsc{Perspicuity} is different ($p=.00033^{***}$) with a small effect size ($r{=}0.42$),
\textsc{Dependability} is different ($p{\leq}.0001^{****}$) with a small effect size ($r{=}0.48$),
\textsc{Stimulation} is different ($p{=}.00047^{***}$) with a small effect size ($r{=}0.40$),
\textsc{Novelty} is different ($p{\leq}.0001^{****}$) with a small effect size ($r{=}0.46$),
\textsc{Trust} is different ($p{\leq}.0001^{****}$) with a small effect size ($r{=}0.47$),
\textsc{Adaptability} is different ($p{\leq}.0001^{****}$) with a medium effect size ($r{=}0.50$),
\textsc{Usefulness} is different ($p{=}.0062^{***}$) with a medium effect size ($r{=}0.31$),
\textsc{Visual aesthetics} are  different ($p{=}.01007^{*}$) with a small effect size ($r{=}0.29$),
\textsc{Intuitive use} is  different ($p{=}.0076^{**}$) with a small effect size ($r{=}0.30$),
\textsc{Trustworthiness} is  different ($p{=}.00028^{***}$) with a small effect size ($r{=}0.41$).
The effect size is interpreted as `medium' more frequently between $S_4$ and $S_{12}$ than between $S_0$ and $S_{12}$ or as `small'. With respect to importance, a significant difference was found between means only for \textsc{Novelty} ($p{=}.15^{*}$) with a large effect size ($r{=}.52$).

\begin{figure}[b]
\centering
\vspace{-16pt}
\includegraphics[width=\linewidth]{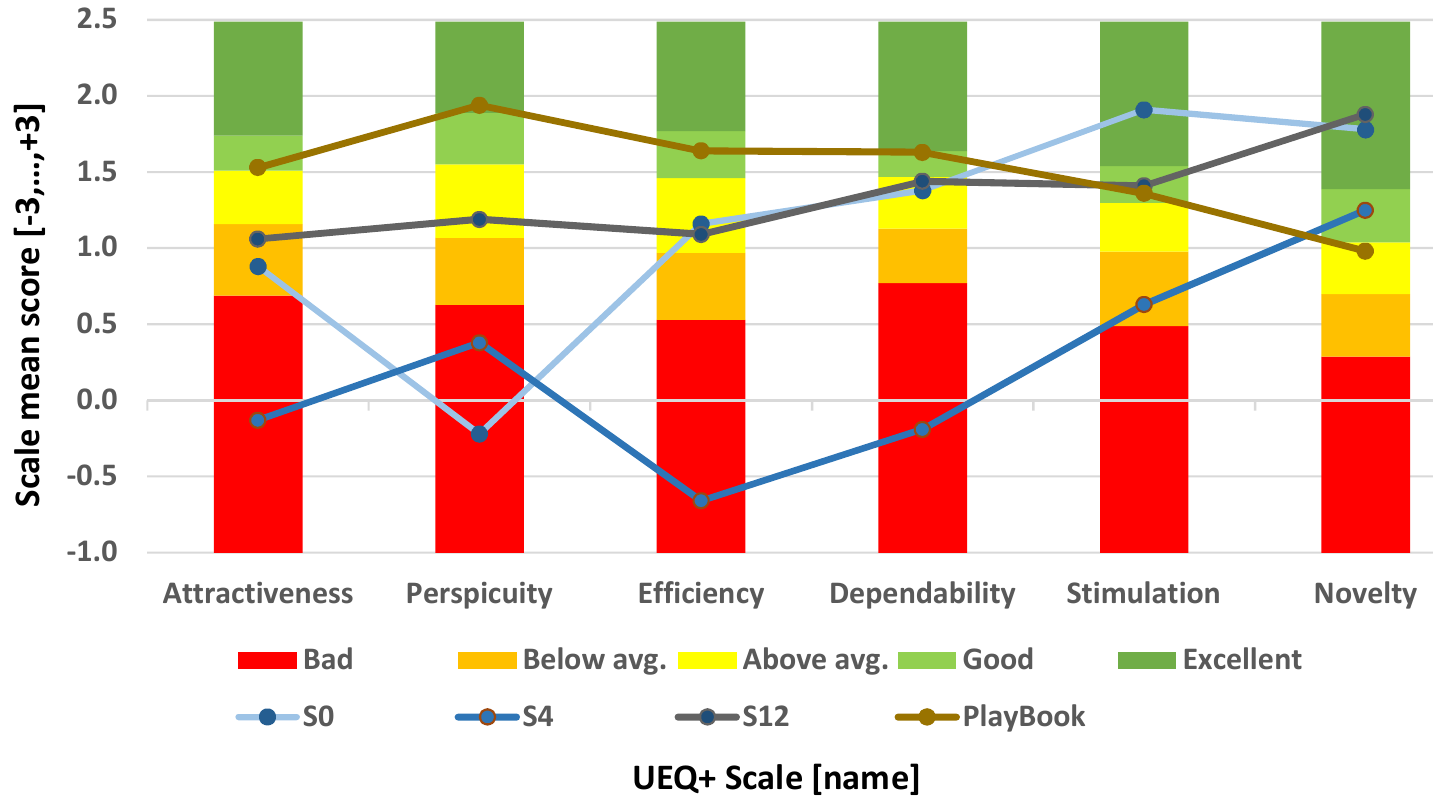}
\vspace{-24pt}
\caption{Benchmarking of UEQ+ scales for each session with respect to PlayBook~\citep{Shelat:2022} in brown. $S_4$,  $S_8$, and $S_{12}$ are represented in light blue, medium blue, and dark blue, respectively.}
\label{fig:benchmarking}
%\vspace{-8pt}
\end{figure}

\subsection{Benchmarking of Scales}
Each UEQ+ scale is typically evaluated as follows: between -0.8 and 0.8 for a neutral evaluation, superior to 0.8 for a positive evaluation, and inferior to -0.08 for a negative evaluation, as explained in the end of Section~\ref{sec:self}. Beyond this universal evaluation, \cite{Schrepp:2017} mentions some more precise intervals for interpreting some of these scales based on a benchmarking obtained by observing the distribution of the values over a large set of evaluated cases. The mean value of each benchmarked scale therefore falls into one of five categories defined as follows~\citep{Schrepp:2017}: \textit{excellent} (among the best 10\% of all cases), \textit{good} (10\% of the cases are better than the evaluated product), \textit{above average} (25\% of the cases are better than the evaluated product), \textit{below average} (50\% of the cases  are better than the evaluated product), and \textit{bad} (the evaluated product is among the worst 25\% of cases). 

\begin{figure*}
\centering
\includegraphics[width=\textwidth]{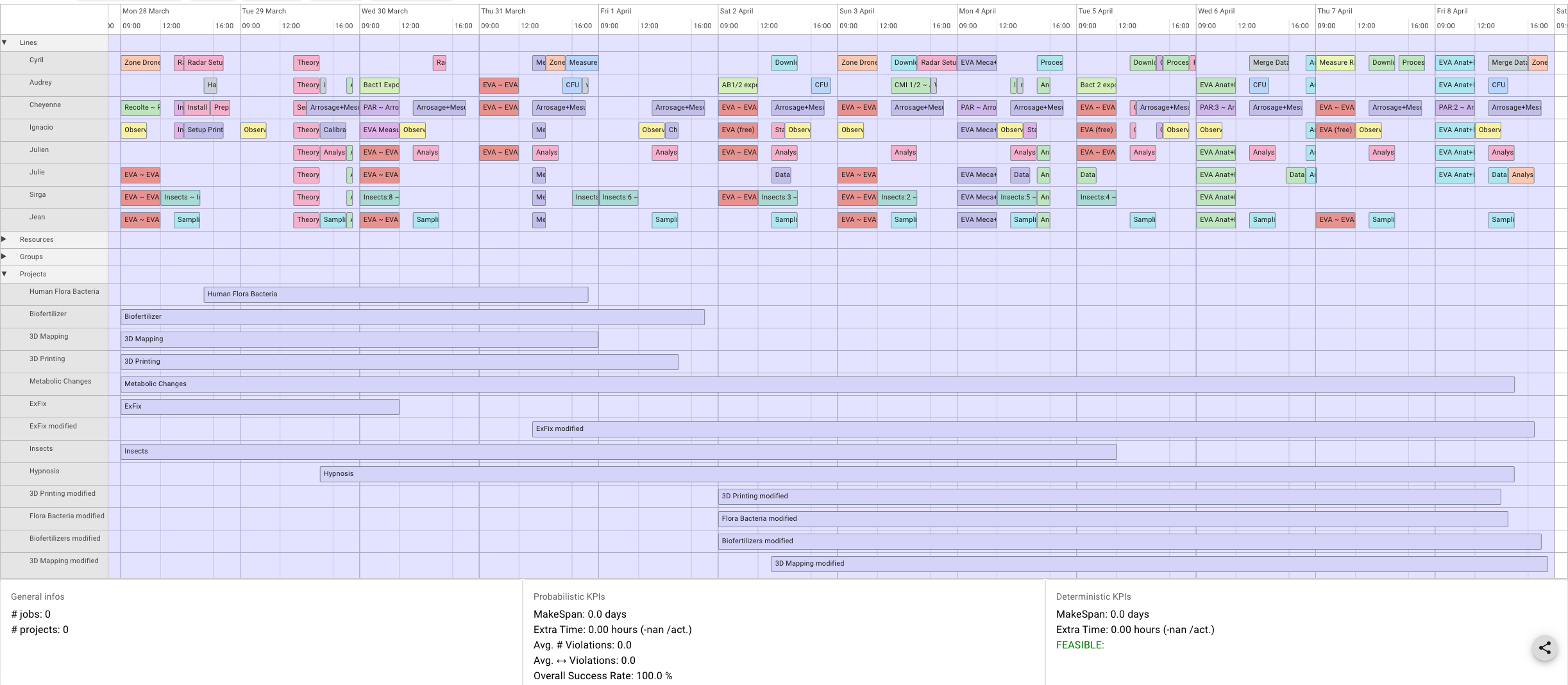}
\caption{Final schedule at the end of the mission (Sol 12). In the end, the astronauts managed to complete all the scientific projects, even though several projects had to be adapted during the course of the operations to fit the field realities.}
\label{fig:planning_sol12}
\end{figure*}

Fig.~\ref{fig:benchmarking} shows the distribution of the benchmarked scales according to the five categories and compares it with PlayBook (\citeauthor{Shelat:2022}, \citeyear{Shelat:2022}), another operations management system, developed and tested by NASA in analogue conditions. Playbook is one of the three components of the Minerva suite (Section \ref{sec:self-scheduling-space}).
Scales for $S_{12}$ are benchmarked as follows: \textsc{Attractiveness} is `excellent' ($M{=}1.06{\geq}0.75$) as well as PlayBook ($M{=}1.53{\geq}0.75$), \textsc{Perspicuity} is `excellent' ($M{=}1.19{\geq}0.6$), \textsc{Efficiency} is `excellent' ($M{=}1.09{\geq}0.72$), \textsc{Dependability} is `excellent' ($M{=}1.44{\geq}0.85$), \textsc{Stimulation} is `excellent' ($M{=}1.41{\geq}0.95$) and slightly above PlayBook ($M{=}1.36{\geq}0.75$), and \textsc{Novelty} is `excellent' ($M{=}1.88{\geq}1.1$) and above PlayBook ($M{=}0.98{\leq}0.75$, interpreted as `good'). 
Overall, Playbook provides significantly better \textsc{Perspicuity}.
Compared to Playbook, \textit{Romie} has additional features, such as the visual modeling framework and the (re)optimization engine. 
These come at the cost of a slightly increased complexity of the system from the user's point of view at discovery stage. After a training period of a couple of weeks, these features no longer affect the  usability as its factors are no longer deteriorated.

\section{Conclusion and Future work}

We study the capability of a crew of analogue astronauts, composed of novice planners, to manage the operational schedule of their mission in an autonomous setup, by using a computer-aided decision system. 
Techniques from human computer interaction (HCI) were exploited to measure and analyse how well the participants succeed at doing so: the astronauts were asked to evaluate their experience of the system and the quality of the computed decisions using UEQ+. 

The results gathered before, and at different stages of the mission, show that the proposed decision system appears as being an adequate approach, from a functional point of view (usefulness), whereas it is perceived as difficult to use by the participants, especially during the first days of the mission. 

Empirical evidence has shown that even provided a strong provisional schedule, rethinking and reshaping all the a priori decisions related to the research projects, to be carried out during the mission, is unavoidable.   
As activities take place, the scientific objectives and constraints must be adapted according to unpredictable events. 
EVAs must be cancelled due to bad weather conditions. The entire project must be adapted to fit the limited duration of the mission. Figure \ref{fig:planning_sol12} shows the planning at the end of the mission. 
Due to the inherent complexity of the underlying combinatorial problem, modifying the schedule by hand is not an option. 
To that extent, the tested decision system includes an artificial intelligence, which proved its usefulness by computing optimised solutions to the scheduling problem, for the astronauts, based on a graphical description of their objectives and constraints. 
The main limitation of the approach lies in the learning time required by the participants to master the system. Future missions will need a more adequate preparation.

\paragraph{Astronaut qualitative impression}
Finally, apart from the quantitative analysis of the questionnaires, the eight astronauts have been asked to describe their impression on the technology, what they liked or disliked: ``\textit{We believe that Rombio is a very useful program for this type of mission, which involves a lot of constraints simultaneously in terms of personnel, time and equipment. Although it was not yet fully functional at the beginning of the mission, it was able to perform a maximum number of pre-programmed activities. The interface could be improved to become more intuitive. Indeed, without adequate prior training, it is very difficult to get used to it, which was relatively energy consuming and time consuming at the beginning. However, as the mission evolved, the program was continually readapted to finally offer us the adequate and appreciated handling for the autonomous management of operations.}''

\paragraph{Acknowledgements} 
We thank the Mars Society, Shannon Rupert and Mission Support personnel, for creating the conditions for realistic Mars analogue sojourns. We thank Agathe Florio and Nathan Gurnet for their helpful statistical feedback. Portions of this work were performed by the Jet Propulsion Laboratory, California Institute of Technology under a contract with the National Aeronautics and Space Administration (80NM0018D0004). Nicolas Burny and Jean Vanderdonckt are supported by the EU EIC Pathfinder-Awareness Inside challenge "\href{http://symbiotik-infovis.eug/}{Symbiotik}" project under Grant no. 101071147.

% \bibliographystyle{elsarticle-harv} 
% \bibliography{bibliography}

%% Bibliography
%% Author year style
\bibliographystyle{jasr-model5-names}
\biboptions{authoryear}
\bibliography{bibliography}

\newpage
\appendix
\section{Evaluation Results Tables}

\begin{table*}[h]
\centering
\caption{Consistency reliability. Cronbach's $\alpha$: $\geq0.9$=excellent (E), $0.9{>}\alpha\geq0.8$=good (G), $0.8{>}\alpha\geq0.7$=acceptable (A), $0.7{>}\alpha\geq0.6$=questionable (Q), $0.6{>}\alpha\geq0.5$=poor (P), $0.5{>}\alpha$=unacceptable (U).}
%\resizebox{\linewidth}{!}{
\begin{tabular}{lrrrr}
\toprule
\multirow{2}{*}{Scale} & \multicolumn{4}{c}{Cronbach's $\alpha$ (interpretation)}  \\
                      \cmidrule{2-5}   
                       & \multicolumn{1}{c}{$S_{0}$}            & \multicolumn{1}{c}{$S_{4}$}           & \multicolumn{1}{c}{$S_{8}$}           & \multicolumn{1}{c}{$S_{12}$}                \\
                     \midrule
 Attractiveness       &   0.62 (Q) &    0.68 (Q) &   0.55 (P)    &   0.91 (E)     \\
 Efficiency           &   0.46 (U) &    -0.10 (U) &   0.70 (A) &   0.52 (P)     \\
 Perspicuity          &   0.95 (E) &    0.56 (P) &   0.48 (U)    &   0.78 (A)      \\
 Dependability        &   0.67 (Q) &    0.75 (A) &   0.72 (A)    &   0.67 (Q)      \\
 Stimulation          &   0.94 (E) &    0.96 (E) &   0.93 (E)    &   0.89 (G)      \\
 Novelty              &   0.89 (G) &    0.63 (Q) &   0.63 (Q)    &   0.63 (Q)      \\
 Trust                &   0.76 (A) &    -0.25 (U) &   0.82 (G)    &   0.88 (G)      \\
 Adaptability         &   0.85 (G) &    0.80 (G) &   0.68 (Q)    &   0.93 (E)      \\
 Usefulness           &   0.69 (Q) &    0.91 (E) &   0.81 (G)   &   0.88 (G)      \\
 Visual Aesthetics    &   0.96 (E) &    0.70 (A) &   0.82 (G)    &   0.76 (A)      \\
 Intuitive Use        &   0.88 (G) &    0.53 (P) &   0.53 (P)    &   0.83 (G)      \\
 Trustworthiness      &   0.76 (A) &    0.55 (P) &   0.77 (A)    &   0.84 (G)      \\
 \midrule
 Mean                 & 0.79 (A) & 0.61 (Q) & 0.70 (A) & 0.80 (G) \\
                       \bottomrule
\end{tabular}
%}
\label{tbl:consistency}
\end{table*}

\begin{table*}[h]
\centering
\caption{Evolution of scale means and their importance rates across sessions in terms of difference of percentage ($M$=mean scale, Imp.=importance rate).}
%\resizebox{\linewidth}{!}{
\begin{tabular}{lrrrrrrrr}
\toprule
\multirow{2}{*}{Scale} & \multicolumn{2}{c}{$S_0 \rightarrow S_4$} & \multicolumn{2}{c}{$S_4 \rightarrow S_8$} & \multicolumn{2}{c}{$S_8 \rightarrow S_{12}$} & \multicolumn{2}{c}{$S_0 \rightarrow S_{12}$} \\
                      \cmidrule{2-3} \cmidrule{4-5} \cmidrule{6-7} \cmidrule{8-9}    
  & \multicolumn{1}{c}{$M$} & \multicolumn{1}{c}{Imp.} & \multicolumn{1}{c}{$M$} & \multicolumn{1}{c}{Imp.}  & \multicolumn{1}{c}{$M$} & \multicolumn{1}{c}{Imp.} &  \multicolumn{1}{c}{$M$} & \multicolumn{1}{c}{Imp.} \\
                      \midrule
 Attractiveness       &   -114\% &   80\%  &   700\%    & -33\% & 42\% &  50\%  &  21\% & 80 \%    \\
 Efficiency           &   -157\% &   19\%  &   257\%    & -11\% &  6\% &  12\%  &  -5\% & 19 \%    \\
 Perspicuity          &    271\% &   13\%  &    75\%    & -24\% & 81\% &  23\%  & 643\% & 7 \%    \\
 Dependability        &   -114\% &    0\%  &   500\%    &   7\% & 92\% &  13\%  &   5\% & 21 \%    \\
 Stimulation          &    -67\% &   17\%  &    30\%    & -14\% & 73\% & 117\%  & -26\% & 117 \%    \\
 Novelty              &    -30\% & -111\%  &    10\%    & 600\% & 36\% & 180\%  &   5\% & 56 \%    \\
 Trust                &    -30\% &    0\%  &    43\%    & -17\% & 70\% & -10\%  &  70\% & -25 \%    \\
 Adaptability         &   -146\% &   15\%  &   369\%    &   0\% &  0\% &   0\%  &  25\% & 15 \%    \\
 Usefulness           &    -67\% &  -18\%  &   110\%    & -11\% & -2\% &   6\%  & -33\% & -23 \%    \\
 Visual Aesthetics    &   -191\% &  -33\%  &    70\%    & 100\% & 41\% &  25\%  & 318\% & 67 \%    \\
 Intuitive Use        &     40\% &  -18\%  &   129\%    & -14\% & 13\% &  33\%  & 260\% & -6 \%    \\
 Trustworthiness      &    -60\% &  -17\%  &   104\%    & -7\%  & 13\% &  14\%  &  -9\% & -11\%    \\
                      \bottomrule
\end{tabular}
%}
\label{tbl:diff_percentage}
\end{table*}

\begin{table*}[]
    \centering
    \caption{Inter-rater agreement. Kendall's $W{\leq}0.2$=low (L), $0.21{\leq}W{\leq}0.4$=fair (F), $0.41{\leq}W{\leq}0.6$=moderate (M), $0.61{\leq}W{\leq}0.8$=high (S), $0.81{\leq}W{\leq}1$=very high (V).}
%        \resizebox{\linewidth}{!}{
\begin{tabular}{lrrrrr}
\toprule
\multirow{2}{*}{Scale} & \multicolumn{4}{c}{Kendall's $W$ ($p$-value, interpretation)}  \\
                       \cmidrule{2-2} \cmidrule{3-3} \cmidrule{4-4} \cmidrule{5-5} \cmidrule{6-6}   
                       & \multicolumn{1}{c}{$S_{0}$}            & \multicolumn{1}{c}{$S_{4}$}           & \multicolumn{1}{c}{$S_{8}$}           & \multicolumn{1}{c}{$S_{12}$}          & \multicolumn{1}{c}{Mean}       \\
   \midrule
 Attractiveness       &   0.14 (0.35, L)  &    0.31 (0.057, F) &   0.094 (0.52, L)    &   0.08 (0.59, L)  & 0.156 (L)   \\
 Efficiency           &   0.098 (0.50, L) &    0.40 (0.021, F) &   0.034 (0.84, L) &   0.023 (0.90, L) &  0.14 (L)   \\
 Perspicuity          &   0.11 (0.43, L)  &    0.0394 (0.81, L) &   0.18 (0.23, L)    &   0.12 (0.40, L) & 0.11 (L)      \\
 Dependability        &   0.041 (0.80, L) &    0.21 (0.15, F) &   0.16 (0.26, L)    &   0.11 (0.44, L) & 0.13 (L)     \\
 Stimulation          &   0.16 (0.26, L)  &    0.077 (0.60, L) &   0.095 (0.51, L)    &   0.023 (0.90, L)  & 0.08 (L)    \\
 Novelty              &   0.042 (0.79, L) &    0.014 (0.95, L) &   0.03 (0.87, L)    &   0.13 (0.37, L) & 0.05 (L)     \\
 Trust                &   0.20 (0.17, L)  &    0.02 (0.92, L) &   0.05 (0.75, L)    &   0.058 (0.71, L) & 0.08 (L)     \\
 Adaptability         &   0.11 (0.43, L)  &    0.058 (0.70, L) &   0.045 (0.78, L)    &   0.056 (0.72, L)  & 0.07 (L)    \\
 Usefulness           &   0.34 (0.041, F)   &  0.25 (0.11, F) &   0.13 (0.36, L)   &   0.26 (0.10, F) & 0.24 (F)     \\
 Visual Aesthetics    &   0.21 (0.16, F)  &    0.15 (0.32, L) &   0.039 (0.81, L)    &   0.10 (0.48, L) & 0.12 (L)     \\
 Intuitive Use        &   0.17 (0.24, L)  &    0.084 (0.57, L) &   0.13 (0.36, L)    &   0.028 (0.88, L) & 0.10 (L)      \\
 Trustworthiness      &   0.077 (0.60, L) &    0.24 (0.12, F) &   0.14 (0.34, L)    &   0.22 (0.14, F) & 0.17 (L)     \\
 \midrule
 Mean & 0.14 (L) & 0.15 (L) & 0.09 (L) & 0.10 (L) \\
 \bottomrule
\end{tabular}
%}
    \label{tbl:reliability}
\end{table*}

\end{document}